\documentclass[10pt,twocolumn,letterpaper]{article}

\usepackage{iccv}
\usepackage{times}
\usepackage{epsfig}
\usepackage{graphicx}
\usepackage{amsmath}
\usepackage{amssymb}
\usepackage{booktabs}
\usepackage{color}
\usepackage{ifthen}
\usepackage{float}
\usepackage{caption}
\usepackage{algorithm}
\usepackage{algorithmic}

\usepackage[pagebackref=true,breaklinks=true,letterpaper=true,colorlinks,bookmarks=false]{hyperref}

\usepackage[capitalize]{cleveref}
\crefname{section}{Sec.}{Secs.}
\Crefname{section}{Section}{Sections}
\Crefname{table}{Table}{Tables}
\crefname{table}{Tab.}{Tabs.}

\iccvfinalcopy 

\def\iccvPaperID{1325} 

\ificcvfinal\fi

\begin{document}

\title{Towards Interactive Image Inpainting via Robust Sketch Refinement}
\author{Chang Liu\textsuperscript{1} \and Shunxin Xu\textsuperscript{1} \and Jialun Peng\textsuperscript{1} \and Kaidong Zhang\textsuperscript{1} \and Dong Liu$^\dagger$\textsuperscript{1}\\
University of Science and Technology of China\textsuperscript{1}\\
{\tt\small \{lc98041, sxu, pjl\}@mail.ustc.edu.cn, hehedazkd@outlook.com, dongeliu@ustc.edu.cn}
}

\twocolumn[{
 \renewcommand\twocolumn[1][]{#1}%
 \maketitle
 \begin{center}
  \centering
  \includegraphics[width=1.0\textwidth]{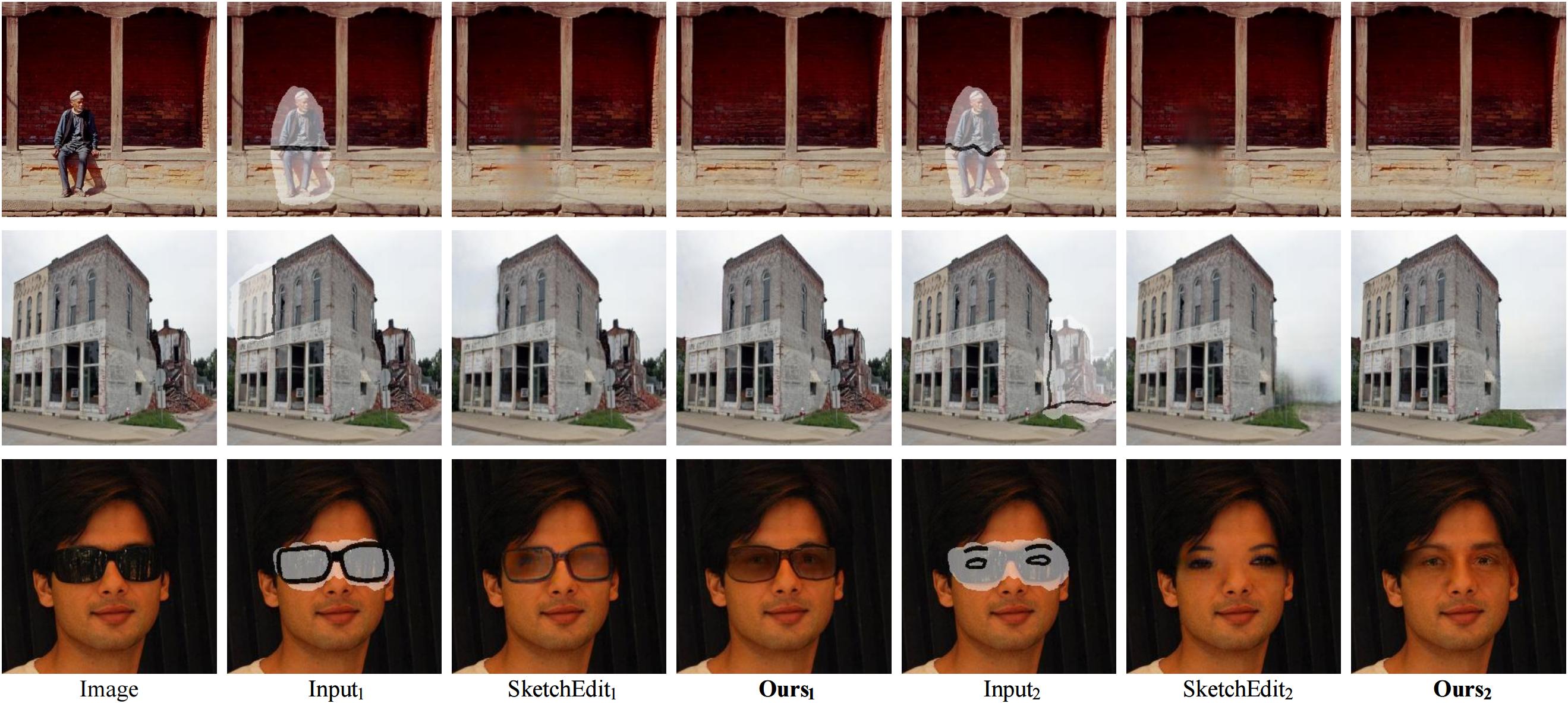}
  \captionof{figure}{\textbf{Interactive image inpainting results of scene editing (top two rows) and face manipulation (the third row)}, produced by SketchEdit \cite{zeng2022sketchedit} and our proposed SketchRefiner. Given user-provided sketches and masks (the second and fifth columns), SketchEdit \cite{zeng2022sketchedit} tries to use the sketches as if they were edges in the missing area, thereby producing noticeable artifacts. In reverse, our method uses the sketches as coarse guiding information, thus reflects user intentions in the produced results and meanwhile tolerates certain randomness of user input. \textbf{Zoom in for best view.} }
  \label{fig1}
 \end{center}}]

\ificcvfinal\thispagestyle{empty}\fi

\let\thefootnote\relax\footnotetext{$^\dagger$ Corresponding author.}

\begin{abstract}
One tough problem of image inpainting is to restore complex structures in the corrupted regions. It motivates interactive image inpainting which leverages additional hints, e.g., sketches, to assist the inpainting process. Sketch is simple and intuitive to end users, but meanwhile has free forms with much randomness. Such randomness may confuse the inpainting models, and incur severe artifacts in completed images. To address this problem, we propose a two-stage image inpainting method termed SketchRefiner. In the first stage, we propose using a cross-correlation loss function to robustly calibrate and refine the user-provided sketches in a coarse-to-fine fashion. In the second stage, we learn to extract informative features from the abstracted sketches in the feature space and modulate the inpainting process. We also propose an algorithm to simulate real sketches automatically and build a test protocol with different applications. Experimental results on public datasets demonstrate that SketchRefiner effectively utilizes sketch information and eliminates the artifacts due to the free-form sketches. Our method consistently outperforms the state-of-the-art ones both qualitatively and quantitatively, meanwhile revealing great potential in real-world applications. Our code and dataset are available at \href{https://github.com/AlonzoLeeeooo/SketchRefiner}{https://github.com/AlonzoLeeeooo/SketchRefiner}.
\end{abstract}

\section{Introduction}

Image inpainting refers to the task of restoring missing pixels in the corrupted regions and simultaneously maintaining a global consistency. Increasing efforts \cite{yu2018free, Nazeri_2019_ICCV, li2019progressive, peng2021generating, liao2021image, guo2021image, suvorov2022resolution, zeng2022sketchedit, li2022mat} have obtained outstanding achievements and put emphasis on numerous practical applications, e.g. face restoration, photo editing, and so on. One challenging problem of image inpainting is to reconstruct complicated structures in the corrupted regions, which yields the proposal of interactive image inpainting leveraging user interactions as guidance. Among numerous interactive physical medias, sketch has stood out as a simple and intuitive way to express creative ideas from users. By simply depicting several lines in the corrupted regions, sketch is able to abstract important elements of an image and convey demands from user straightforwardly.

However, existing image inpainting methods are still limited while handling sketch-like inputs. Among traditional methods, there exist enlightening ideas of bringing user interactions as structural auxiliary \cite{sun2005image, sun2011structure}. But these methods are heuristic and fail to recover semantically reasonable information in the corrupted areas. As learning-based methods \cite{li2019progressive, peng2021generating, liao2021image, guo2021image, suvorov2022resolution, li2022mat} gradually become dominant in the field of image inpainting, edge-based image inpainting methods \cite{yu2018free, Nazeri_2019_ICCV, xiong2019foreground, guo2021image, dong2022incremental} and sketch-based image editing methods \cite{jo2019sc, yang2020deep, liu2021deflocnet, zeng2022sketchedit} obtain improved performance with the assistance of edges or contours. Nevertheless, they lack consideration of the characteristics of the sketch domain and request pixel-wise preciseness in the mask regions. As is shown in Fig.~\ref{fig1}, such sub-optimal solutions cause unnatural results with artifacts and restrict the real-world applications of sketches. Although such limitaions have raised numerous researches \cite{liu2019sketchgan, li2019photo, 10.1145/3528223.3530068, chan2022learning} for sketch refinement or completion, the gap between sketch-like inputs and image inpainting models upon real-world applications still exists, and the art of utilizing sketch interactions needs to be re-investigated.

In this paper, we attempt to overcome the aforementioned limitations with our proposed method, named SketchRefiner. SketchRefiner is a two-stage system handling the tasks of sketch refinement and image inpainting sequentially. It is capable of calibrating user-drawn sketches and modulating the inpainting process with the sketch guidance. In our SketchRefiner, we design a novel sketch refinement network and propose the use of cross-correlation loss to boost its performance. Moreover, we extract informative features from the refined sketches and aggregate them into the inpainting network as modulation. The model is then able to produce the inpainted results according to the sketch inputs. The two-stage architecture of SketchRefiner is highly interpretable and reveals great scalability to other inpainting methods. Furthermore, we propose an algorithm to simulate real sketches automatically. We observe the absence of sketch-based testing benchmarks, and build a test protocol to highlight the simplicity and effectiveness of sketches in real-world applications. Experiments upon ImageNet \cite{russakovsky2015imagenet}, CelebA-HQ \cite{karras2018progressive}, Places \cite{zhou2017places} datasets and the proposed test protocol demonstrate the guaranteed performance of SketchRefiner over other state-of-the-art competitors. Generally speaking, our contributions are three-fold:

\begin{itemize}
\setlength{\itemsep}{0pt}
\setlength{\parsep}{0pt}
\setlength{\parskip}{0pt}
\item We re-investigate the challenges of sketch-based interactive image inpainting, by proposing a two-stage system called SketchRefiner to bridge the existing gap between sketch-like inputs and image inpainting models.
\item Targeting the characteristics of the sketch domain, we propose a sketch refinement network with a novel loss function to robustly calibrate the user-provided sketches. Further, we learn to extract informative features from sketches and aggregate them as modulation in our inpainting network.
\item We propose a sketch simulation algorithm and establish a sketch-based test protocol for real-world applications. Experiments on public datasets and the proposed testing benchmark demonstrate the effectiveness of our method against state-of-the-art competitors.
\end{itemize}

\begin{figure*}[t!]
\centering
\includegraphics[width=1.0\textwidth]{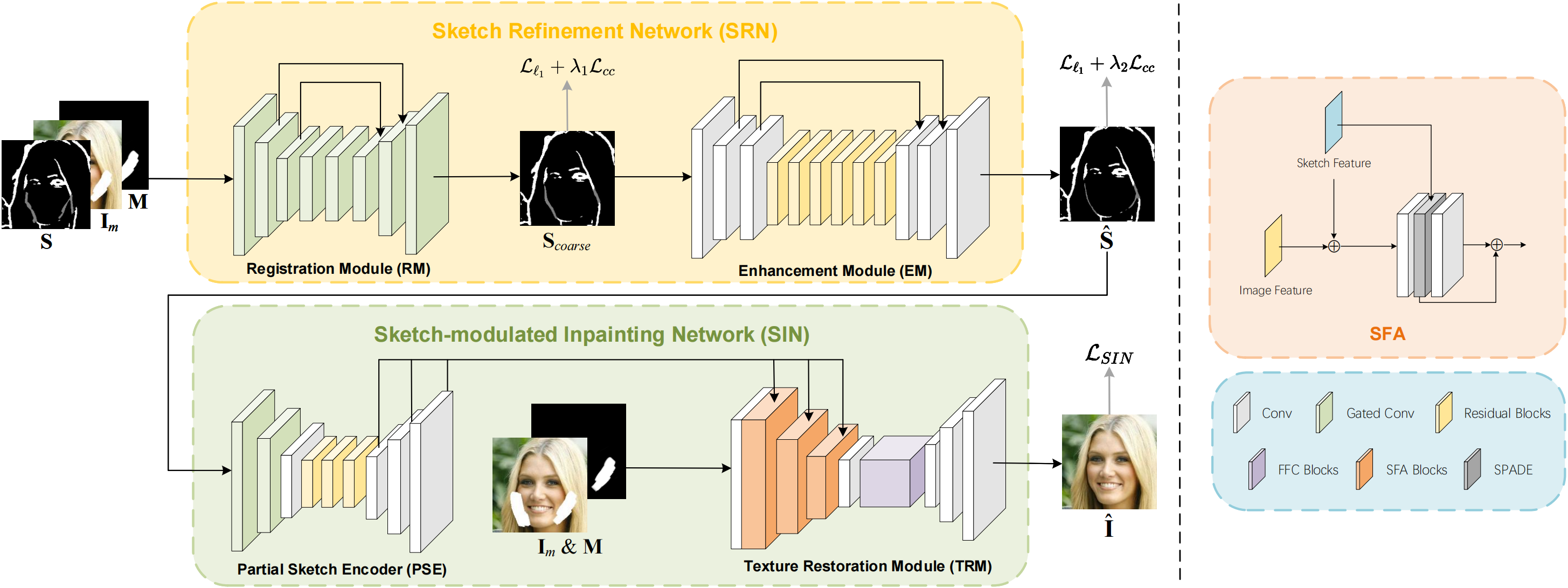} 
\caption{Overview of SketchRefiner (left) and Sketch Feature Aggregation (SFA) block. SketchRefiner consists of Sketch Refinement Network (SRN) and Sketch-modulated Inpainting Network (SIN). To address the \emph{misalignment} of sketches, we first send sketch $\textbf{S}$, corrupted image $\textbf{I}_m$ and mask $\textbf{M}$ into the Registration Module (RM), which produces a coarsely refined sketch $\textbf{S}_{coarse}$. Then, Enhancement Module (EM) further refines the structural \emph{incoherence} in $\textbf{S}_{coarse}$ and yields our final refined sketch $\hat{\textbf{S}}$. Here, $\lambda_1$ and $\lambda_2$ are loss weighing hyperparameters. In SIN, Partial Sketch Encoder (PSE) first learns to extract features from $\hat{\textbf{S}}$. Then, Texture Restoration Module (TRM) merges the features as modulation with Sketch Feature Aggregation (SFA) blocks. Eventually, SIN restores the missing texture and produces the inpainted result $\hat{\textbf{I}}$.}
\label{fig2}
\end{figure*}

\section{Related Work}
\noindent\textbf{Image Inpainting.} Recently, learning-based methods become dominant in the field of image inpainting. Vanilla image inpainting methods, consisting of deterministic \cite{pathak2016context, iizuka2017globally, liu2018image, yu2018free, liu2020rethinking, yu2020region, peng2021generating, liu2021pd, wang2021parallel, yu2021wavefill, suvorov2022resolution, li2022mat, zheng2022bridging} and pluralistic ones \cite{zheng2019pluralistic, zhao2020uctgan, wan2021high, wan2021high, peng2021generating, lugmayr2022repaint, li2022mat}, aim to predict the missing pixels with only information from the valid regions. Since such solutions are often incapable of handling challenging cases with complex structures, edge-based image inpainting methods \cite{yu2018free, Nazeri_2019_ICCV, li2019progressive, xiong2019foreground, liao2021image, guo2021image, cao2021learning} are thus motivated. These methods manage to recover the corrupted images with assistance of estimated edges, and reveal improved performance. However, they request pixel-wise preciseness of the provided guidance and suffer greatly from the existing gap between user-drawn sketches and image inpainting models, which become less practical in real-world applications.

\noindent\textbf{Sketch-based Image Editing.} Sketch guidance has been explored by a number of researches \cite{sun2005image, yu2018free, jo2019sc, yang2020deep, liu2021deflocnet, zeng2022sketchedit} for its simplicity and effectiveness to express user intuition. In order to tackle the absence of large-scale sketch-annotated datasets, these methods use detected edges \cite{xie2015holistically, he2020bdcn} as alternatives during training. However, these solutions lack consideration of the characteristics of the sketch domain, and often lead to sub-optimal results similar to the cases in Fig.~\ref{fig1}. Therefore, the art of adapting current methods to sketch interactions in real-world circumstances needs to be re-investigated.

\noindent\textbf{Sketch Refinement and Completion.} Early researches \cite{sangkloy2016sketchy, eitz2012hdhso, liu2019sketchgan, li2019photo, 10.1145/3528223.3530068, chan2022learning} have proven that sketches are closely related to edges, both of which are visually closed outlines of objects. The major difference between sketches and edges is that edges are pixel-wise corresponding while sketches are more diversified and abstract. Although a number of efforts \cite{ren2005scale, ming2012connected, liu2019sketchgan, Nazeri_2019_ICCV, xiong2019foreground, dong2022incremental} have been motivated, they fail to handle user-drawn sketches in real cases due to a lack of full consideration of other differences between sketches and edges, e.g. \emph{misalignment} and \emph{incoherence}.


\section{How to Use Sketches for Inpainting?}

In this section, we first analysis the advantage and characteristics of sketches compared to edges. We distinguish sketches from edges from three major differences: \emph{misaligned}, \emph{incoherent} and \emph{abstract}. Furthermore, we illustrate our solution of how to simulate sketches for inpainting targeting these characteristics.

\begin{figure}[t!]
\centering
\includegraphics[width=1.0\linewidth]{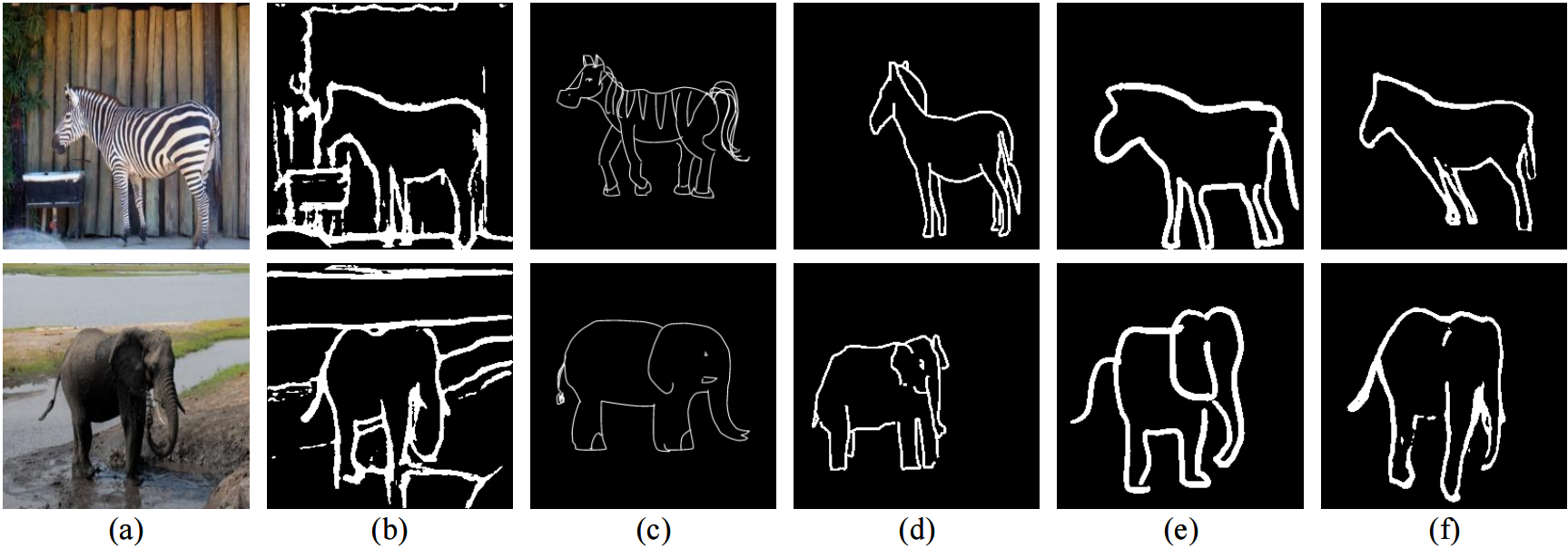} 
\caption{Visualization of edges and different sketches. (a) Image, (b) detected edge \cite{he2020bdcn}, (c) sketch provided in TU-Berlin \cite{eitz2012hdhso}, (d) sketch provided in Sketchy \cite{sangkloy2016sketchy}, (e) user-drawn sketch in our conducted user study, and (f) simulated sketch with SSA (Algorithm~\ref{SSA}). \textbf{Zoom in for best view.}}
\label{DeformedSketches}
\end{figure}

\noindent \textbf{Sketches vs. Edges.} Recent methods \cite{Nazeri_2019_ICCV, li2019progressive, xiong2019foreground, liao2021image, guo2021image, cao2021learning} have revealed performance gain in image inpainting with the guidance of estimated edges. However, predicted edges are always deterministic, and fail to convey user demands in the generated results. In reverse, sketch is more intuitive, and shows more advantages in real applications to express user intuition. We distinguish sketches from edges according to the process of how users sketch from images. Fig.~\ref{DeformedSketches} shows visualization of edges and different sketches \cite{sangkloy2016sketchy, eitz2012hdhso}. First, sketches are always spatially \emph{misaligned} with the original images compared to the pixel-wise corresponding edges. It is tough for untrained users to draw the exactly aligned outlines of an image. Second, sketches are often structurally \emph{incoherent}. User-drawn sketches could be decomposed into a sequence of strokes, where incoherence and breakpoints could occur between each individual stroke. But it is easy for edges to be structurally consistent due to the continuity of images. Third, sketches are more \emph{abstract} than edges, where users first understand the general content, and then sketch out their conception in a semantic level. Thus, it is tough for neural networks to ``understand'' sketches in the same pixel space as edges. 

\noindent \textbf{Sketch Simulation Algorithm.} One long-existing bottleneck of sketch-based image inpainting is the absence of sketch-image datasets for training. The issue motivates numerous solutions \cite{yu2018free, jo2019sc, yang2020deep, liu2021deflocnet, zeng2022sketchedit} to simulate sketches with detected edges \cite{xie2015holistically, he2020bdcn}. However, their solutions fail to distinguish sketches from edges, and lead to sub-optimal results (Fig.~\ref{fig1}). Targeting the characteristics of the sketch domain and the data insufficiency problem, we provide our solution, called Sketch Simulation Algorithm (SSA). We warp detected edges \cite{he2020bdcn} and disrupt their pixel-wise correspondence to simulate the \emph{misalignment} of sketches. Besides, we compose the warped sketches with edges in the valid regions, which imitates \emph{incoherence} of sketches and benefit the process of sketch refinement. To simulate sketch in a more \emph{abstract} level, we detect the edge \cite{he2020bdcn} from the salient parts \cite{qin2020u2} of an image instead of ground truth. Practically, we first detect the foreground $\textbf{I}_f$ and its corresponding edge $\textbf{E}_f$ with a saliency detection model $\mathbb{E}_{SD}$ \cite{qin2020u2} and a edge detection model $\mathbb{E}_{ED}$ \cite{he2020bdcn}. Then, we randomly sample a set of parameters fitting a Gaussian distribution, denoted as $\theta$. We fix a Gaussian Filter (GF) termed $\mathbb{F}_\theta$ with $\theta$ as its weights, where $\mathbb{F}$ is implemented via a convolution layer with a kernel size of $k$. We pass a random Gaussian signal $x$ through $\mathbb{F}_\theta$, and yield a warping map $\phi$ with control of a deformation magnitude $\psi$. Eventually, we apply $\phi$ upon $\textbf{E}_f$ and obtain the simulated sketch $\textbf{S}$. Fig.~\ref{DeformedSketches} shows the visualization of our simulated sketches compared to edges, sketches provided in existing sketch datasets \cite{eitz2012hdhso, sangkloy2016sketchy}, and user-drawn sketches in our conducted user studies. We depict the whole process of SSA in the following algorithm.

\begin{algorithm}
	\renewcommand{\algorithmicrequire}{\textbf{Input:}}
	\renewcommand{\algorithmicensure}{\textbf{Output:}}
	\caption{Pseudo code of SSA}
	\begin{algorithmic}[1]
	\label{SSA}
		\REQUIRE groud truth $\textbf{I}$, foreground image $\textbf{I}_f$, binary mask $\textbf{M}$, image resolution $N$, saliency detection model $\mathbb{E}_{SD}$, edge detection model $\mathbb{E}_{ED}$
		\ENSURE sketch $\textbf{S}$
        \STATE foreground $\textbf{I}_f \leftarrow \mathbb{E}_{SD}(\textbf{I})$
        \STATE foreground edge $\textbf{E}_f \leftarrow \mathbb{E}_{ED}(\textbf{I}_f)$
		\STATE Gaussian weights $\theta \sim N(0,1)$ in size of $1\times 1\times k\times k$\\
		\STATE GF $\mathbb{F}_\theta \leftarrow$ \texttt{nn}.\texttt{Conv2d(1,1,kernel\_size=$k$)}
		\STATE random Gaussian signal $x \sim N(0,1)$ in size of $B\times C\times H\times W$\\
	    \STATE deformation magnitude $\psi \sim U[\psi_{min}, \psi_{max}]$\\
			   warping map $\phi \leftarrow \psi \odot \mathbb{F}_{\theta}(x)$\\
		\STATE $\textbf{S} \leftarrow$ \texttt{F.grid\_sample$(\textbf{E}_f$,$\phi)$}\\
			   $\textbf{S} \leftarrow \textbf{S} \odot \textbf{M} + \textbf{E} \  \odot$ $(1-\textbf{M})$
	\end{algorithmic}
\end{algorithm}

\section{Proposed Method}

In this section, we illustrate how we manage to calibrate sketch-like inputs and how we modulate the inpainting process with sketches. Fig.~\ref{fig2} shows the pipeline of the proposed method. We divide the pipeline into sketch refinement and sketch-modulated inpainting. During sketch refinement, we first propose a novel Sketch Refinement Network (SRN) to tackle the \emph{misalignment} and \emph{incoherence} of sketches. We introduce Cross-Correlation (CC) Loss to provide a region-level constrain for sketch refinement. During inpainting, we propose Partial Sketch Encoder (PSE) to learn to extract informative features from sketches in the feature space. In Texture Restoration Module (TRM), we use Sketch Feature Aggregation (SFA) blocks to fuse the extracted features as modulation of the inpainting process.

\subsection{Sketch Refinement Network}

The goal of SRN is to calibrate the \emph{misalignment} and \emph{incoherence} of sketches. First of all, we require a precise and well-performed signal as the supervision of SRN. Edge maps \cite{xie2015holistically, he2020bdcn} are important features of images to reflect their pixel-wise precise outlines, and bring improvement to the inpainting process \cite{yu2018free, Nazeri_2019_ICCV, li2019progressive, xiong2019foreground, liao2021image, guo2021image, cao2021learning}. To this end, we set the foreground edge $\textbf{E}_f$ as the ground truth of SRN, expecting to standardize sketches to the greatest extent. In SRN, we manage to tackle the \emph{misalignment} and \emph{incoherence} of sketches with Registration Module (RM) and Enhancement Module (EM) correspondingly. We first send concatenation of masked image $\textbf{I}_{m}$, mask $\textbf{M}$ and sketch $\textbf{S}$ into RM. We implement gated convolution blocks \cite{yu2018free} in RM to selectively utilize image information in the valid region. In RM, we expect to diminish those \emph{misalignment} of sketches and justify them to reasonable positions. Then, EM further strengthens the structural \emph{coherence} of $\textbf{S}_{coarse}$. We send $\textbf{S}_{coarse}$ into EM and use vanilla convolutions in all layers. SRN eventually calibrates $\textbf{S}$ into the refined sketch $\hat{\textbf{S}}$, which is formulated as:
\begin{equation}
    \textbf{S}_{coarse} = \mathrm{RM}(\textbf{I}_{m}, \textbf{M}, \textbf{S}),\ \  \hat{\textbf{S}} = \mathrm{EM}(\textbf{S}_{coarse}).\\
\end{equation}

\noindent \textbf{Cross-Correlation Loss.} Since \emph{misalignment} and \emph{incoherence} occur as region-level problems, it is essential to penalize intensified differences in regions of the refined sketches in SRN. Thus, we propose the use of Cross-Correlation (CC) Loss based on \cite{balakrishnan2019voxelmorph}. Cross-correlation loss computes local means over the refined sketch $\hat{\textbf{S}}$ and ground truth edge $\textbf{E}$ within a sliding grid in size of $n^2$. We compute the local means over $\hat{\textbf{S}}$ and $\textbf{E}$ using a fixed convolution layer with all one weights. Since a higher cross-correlation value indicates a better alignment between $\hat{\textbf{S}}$ and $\textbf{E}$, we set the cross-correlation value negative, yielding the loss function as:\par
\begin{small}
\begin{equation}
    \mathcal{L}_{cc}=-CC(\hat{\textbf{S}}, \textbf{E}) = 
    -\sum_{p\in\Omega} \frac{\left(\sum_{p_i}\Delta\hat{\textbf{S}}_{p_i}\Delta\textbf{E}_{p_i}
    \right)^2}
    {\left(\sum_{p_i}\Delta\hat{\textbf{S}}_{p_i}\right)
    \left(\sum_{p_i}\Delta\textbf{E}_{p_i}\right)},
\label{eq3}
\end{equation}
\end{small}

where $\Omega$ denotes the whole local region. $p_i$ and $p$ are sliding grids in size of $n^2$. $p$ is first fixed and $p_i$ iterates around $p$. Then $p$ iterates over $\Omega$. Both $p_i$ and $p$ iterate with a stride of $1$. For every fixed $p$, we define difference map $\Delta\textbf{A}$ as L1 norm between $\textbf{A}_{p_i}$ and $\textbf{A}_{p}$, formed as $\Delta\textbf{A}_{p_i} = \Vert\textbf{A}_{p_i}-\textbf{A}_{p}\Vert_1$. Here, $\textbf{A}_{p_i}$ and $\textbf{A}_{p}$ represent the local means within the sliding grid $p_i$ and the fixed one $p$, respectively. $\Delta\hat{\textbf{S}}_{p_i}$ and $\Delta\textbf{E}_{p_i}$ represent difference maps of refined sketch $\hat{\textbf{S}}$ and the ground truth $\textbf{E}$.

We set loss functions of RM and EM as different portions of combinations between pixel-wise constrain $\mathcal{L}_{\ell_1}$ and region-wise constrain $\mathcal{L}_{cc}$, formulated as:
\begin{equation}
\mathcal{L}_{RM} = \mathcal{L}_{\ell_1} + \lambda_{1}\mathcal{L}_{cc},\  \mathcal{L}_{EM} = \mathcal{L}_{\ell_1} + \lambda_2\mathcal{L}_{cc},\\
\end{equation}

where the pixel-wise constrain is written as $\mathcal{L}_{\ell_1} = \Vert\hat{\textbf{S}} - \textbf{E}\Vert_1$. Since RM and EM handle \emph{misalignment} and \emph{incoherence} of sketches correspondingly, the term of region-wise constrain $\lambda_1\mathcal{L}_{cc}$ in $\mathcal{L}_{RM}$ should have a larger value than $\lambda_2\mathcal{L}_{cc}$ in $\mathcal{L}_{EM}$, which results in a larger $\lambda_2$ than $\lambda_1$. Empirically, we set $\lambda_1 = 0.4$ and $\lambda_2 = 0.9$ for best performance.

\begin{table*}[t]
    \centering
    \begin{tabular}{lccc}
     \toprule
     &CelebA-HQ &Places2 &ImageNet \\
     Method     & PSNR$^\uparrow$ / SSIM$^\uparrow$ / FID$^\downarrow$ & PSNR$^\uparrow$ / SSIM$^\uparrow$ / FID$^\downarrow$ & PSNR$^\uparrow$ / SSIM$^\uparrow$ / FID$^\downarrow$\\
     \midrule
     DeepFill-v2\cite{yu2018free}     & 22.06 \ / \ 0.788 \ / \ 26.78 \ & \ 24.42 \ / \ \textcolor{blue}{0.872} \ / \ 6.71 \ & \ 20.46 \ / \ 0.745 \ / \ 22.34 \\
     DeepPS\cite{yang2020deep}     & 20.59 \ / \ 0.739 \ / \ 72.74 \ & \ \ \ \ \ - \ \ \ \ \ / \ \ \ \ \ - \ \ \ \ \ / \ \ \ \ - \ \ \ \ & \ \ \ \ \ - \ \ \ \ \ / \ \ \ \ \ - \ \ \ \ \ / \ \ \ \ - \ \ \ \ \\
     LaMa\cite{suvorov2022resolution} & \textcolor{blue}{23.48} \ / \ \textcolor{blue}{0.825} \ / \ 19.13 \ & \ 24.17 \ / \ 0.871 \ / \ 6.57 \ & \ \textcolor{blue}{21.25} \ / \ \textcolor{blue}{0.752} \ / \ \textcolor{blue}{14.31} \\
     SketchEdit\cite{zeng2022sketchedit}     & 19.32 \ / \ 0.737 \ / \ 55.64 \ & \ 20.39 \ / \ 0.713 \ / \ 19.3 \ & \ \ \ \ \ - \ \ \ \ \ / \ \ \ \ \ - \ \ \ \ \ / \ \ \ \ - \ \ \ \ \\
     SketchEdit*\cite{zeng2022sketchedit}     & 20.21 \ / \ 0.759 \ / \ 41.28 \ & \ 22.08 \ / \ 0.806 \ / \ 11.1 \ & \ \ \ \ \ - \ \ \ \ \ / \ \ \ \ \ - \ \ \ \ \ / \ \ \ \ - \ \ \ \ \\
     ZITS\cite{dong2022incremental}     & 22.79 \ / \ 0.812 \ / \ \textcolor{blue}{18.79} \ & \ \textcolor{blue}{24.50} \ / \ 0.871 \ / \ \textcolor{blue}{6.20} \ & \ 20.98 \ / \ 0.743 \ / \ 17.71 \\
     \textbf{Ours}     & \textcolor{red}{23.90} \ / \ \textcolor{red}{0.831} \ / \ \textcolor{red}{16.32} \ & \ \textcolor{red}{24.99} \ / \ \textcolor{red}{0.877} \ / \ \textcolor{red}{5.53} \ & \ \textcolor{red}{22.16} \ / \ \textcolor{red}{0.769} \ / \ \textcolor{red}{9.531} \\
     \bottomrule
\end{tabular}
    \caption{Quantitative results with synthetic samples on CelebA-HQ \cite{karras2018progressive}, Places \cite{zhou2017places} and ImageNet \cite{russakovsky2015imagenet}, compared to DeepFill-v2 \cite{yu2018free}, DeepPS \cite{yang2020deep}, LaMa \cite{suvorov2022resolution}, SketchEdit \cite{zeng2022sketchedit}, and ZITS \cite{dong2022incremental}. Here, the \textcolor{red}{best} and \textcolor{blue}{second best} results are in red and blue. SketchEdit represents the results using predicted masks \cite{zeng2022sketchedit} and SketchEdit* represents the results using the same masks as others. "-" stands for unavailable results because the corresponding methods did not perform experiments on that dataset.}
    \label{table1}
\end{table*}

\begin{table}[t!]
    \centering
    \setlength{\tabcolsep}{0.8mm}
    \small
    \scalebox{0.85}{
    \begin{tabular}{lcc}
     \toprule
     & \  Face Editing \  & \ Object Removal \  \\
     \  Method \  & IS$^\uparrow$ \ /  \ SL\texttt{\_1}$^\downarrow$\  / SL\texttt{\_2}$^\downarrow$ \ & IS$^\uparrow$ \ /  \ SL\texttt{\_1}$^\downarrow$\  / SL\texttt{\_2}$^\downarrow$ \ \\
	 \midrule     
     DeepPS \cite{yang2020deep} & \ 2.973 \ / \ 0.0039 \ / \ 0.0154 \ & \ - \ \ \ / \ \ \ \ \ \ - \ \ \ \ \ \ / \ \ \ - \ \ \\
     DeepFill-v2 \cite{yu2018free} & \ 3.149 \ / \ \textcolor{blue}{0.0018} \ / \ \textcolor{blue}{0.0078} \ & \ 4.516 \ / \ 0.0081 \ / \ 0.0393 \ \\
     LaMa \cite{suvorov2022resolution} & \ \textcolor{blue}{3.287} \ / \ 0.0023 \ / \ 0.0107 \  & \ 4.692 \ / \ 0.0078 \ / \ 0.0381 \ \\
     SketchEdit \cite{zeng2022sketchedit} & \ 3.058 \ / \ 0.0033 \ / \ 0.0123 \  & \ 4.669 \ / \ 0.0089 \ / \ 0.0479 \ \\
     ZITS \cite{dong2022incremental} & \ 3.268 \ / \ 0.0019 \ / \ 0.0098 \ & \ \textcolor{blue}{4.711} \ / \ \textcolor{blue}{0.0074} \ / \ \textcolor{blue}{0.0369} \ \\
     \textbf{Ours} & \ \textcolor{red}{3.339} \ / \ \textcolor{red}{0.0013} \ / \ \textcolor{red}{0.0076} \ & \ \textcolor{red}{4.840} \ / \ \textcolor{red}{0.0067} \ / \ \textcolor{red}{0.0354} \ \\
     \bottomrule
     \end{tabular}}
    \caption{Quantitative results on our proposed test protocol, compared to DeepPS \cite{yang2020deep}, DeepFill-v2 \cite{yu2018free}, LaMa \cite{suvorov2022resolution}, SketchEdit \cite{zeng2022sketchedit}, and ZITS \cite{dong2022incremental}. Here, the \textcolor{red}{best} and \textcolor{blue}{second best} results are in red and blue.}
    \label{table2}
\end{table}

\subsection{Sketch-modulated Image Inpainting}
 

\noindent \textbf{Partial Sketch Encoder.} To introduce additional guidance into the inpainting process, e.g., edge, a popular solution in previous methods \cite{Nazeri_2019_ICCV, yu2018free, cao2021learning} is concatenating it as an additional channel of the network. However, since sketches are human \emph{abstraction} of images, it would be tough for neural networks to ``understand'' them in the same pixel space as edges. To this end, we propose Partial Sketch Encoder (PSE) to learn to encode sketches onto the feature space and extract informative features from them. In PSE, we use gated convolution \cite{yu2018free} in the first two layers following \cite{dong2022incremental}. We select a series of coarse-to-fine features from the last four layers in PSE, expecting to extract multidimensional information in the feature space. The features are then aggregated to TRM to modulate the inpainting process.

\noindent \textbf{Sketch Feature Aggregation.} Considering that the extracted features by PSE are a series of multidimensional features, it is unsuitable to merge them directly into the original layers in \cite{suvorov2022resolution}. Therefore, we propose SFA blocks to learn to aggregate these features. Fig.~\ref{fig2} shows the detailed design of the SFA block. In SFA blocks, we first project the encoded features onto an embedding space. Then, we convolute them to two modulation tensors $\gamma$ and $\beta$. The modulation tensors are merged to the image features by element-wise adding and multiplying $\gamma$ and $\beta$ to the activation layers. We also set $\gamma$ and $\beta$ learnable, aiming to learn the feature aggregation process. Practically, we use \cite{park2019semantic} to linearly transform the encoded features onto the embedding space and use Batch Normalization \cite{ioffe2015batch} in our SFA blocks.


\noindent \textbf{Texture Restoration Module.} We conduct TRM as our inpainting network based on \cite{suvorov2022resolution} for its well performance and efficiency. Further, we implement our aforementioned SFA blocks in the encoder part of TRM.

\noindent \textbf{Loss Functions.} We follow the loss function setting in \cite{suvorov2022resolution}, consisting of pixel-wise reconstruction loss $\mathcal{L}_{\ell_1}$, adversarial loss $\mathcal{L}_{adv}$ \cite{gulrajani2017improved}, feature matching loss $\mathcal{L}_{fm}$ \cite{wang2018high}, and High Receptive Field (HRF) perceptual loss $\mathcal{L}_{HRF}$ \cite{suvorov2022resolution}. The overall loss function $\mathcal{L}_{SIN}$ is written as:\par
\begin{small}
\begin{equation}
\mathcal{L}_{SIN} = \lambda_{\ell_1}\mathcal{L}_{\ell_1} + \lambda_{adv}\mathcal{L}_{adv} + \lambda_{fm}\mathcal{L}_{fm} + \lambda_{HRF}\mathcal{L}_{HRF},
\end{equation}
\end{small}
where $\lambda_{\ell_1}$, $\lambda_{adv}$, $\lambda_{fm}$, $\lambda_{HRF}$ are hyperparameters. $\mathcal{L}_{adv}$ is the adversarial loss with gradient pernalty \cite{gulrajani2017improved}, and $\mathcal{L}_{fm}$ \cite{wang2018high} is used to stabilize the training process of GAN. $\mathcal{L}_{HRF}$ \cite{suvorov2022resolution} calculates the distance of extracted features from a backbone network with higher receptive field. $\mathcal{L}_{HRF}$ provides a high-level constrain between the sketch-modulated inpainted result and the ground truth in SIN.

\section{Experiments}
\label{sec4}

In this section, we first introduce our proposal of a sketch-based testing benchmark. Then, we demonstrate the performance of the proposed technique, compared to state-of-the-art image inpainting and image editing methods. We evaluate the compared methods using quantitative metrics with both synthetic samples and our proposed benchmark. Then, we ablate the effect of sketch refinement, components of SRN, as well as components of SketchRefiner. We introduce the implementation details, the training strategies, and our model scalability in our supplementary materials.


\subsection{Experimental Setup}

\noindent\textbf{Sketch-Based Test Protocol.} We observe the absence of real-world testing benchmark in the field of sketch-based image inpainting. Therefore, we build a sketch-based test protocol to evaluate the performance of models in real-world circumstances. The test protocol concentrates on both human face manipulation and scene-level editing. We first collect a number of face and scene-level images from the Internet. Then, users are requested to imagine how they would edit such images and draw corresponding sketch and mask. Eventually, we observe that the collected test protocol is able to cover a series of real-world applications, e.g., object removal, face manipulation and so on. Both sketches and masks are produced using the application SketchBook on iPad with an apple pencil. And we only preserve the sketch parts in the mask region for evaluation.

\noindent\textbf{Datasets and Metrics.} We implement our experiments on ImageNet \cite{russakovsky2015imagenet}, CelebA-HQ \cite{li2019progressive} and Places \cite{zhou2017places}. For evaluation, we randomly collect 10,000 unseen images from test sets of ImageNet \cite{russakovsky2015imagenet}, Places2 \cite{zhou2017places} and 1,000 unseen images from CelebA-HQ \cite{karras2018progressive}. We follow the mask setting in \cite{yu2018free} and use processed sketches with our SSA. For evaluation on synthetic samples, we use common metrics including PSNR, SSIM and FID \cite{heusel2017gans} for quantitative evaluation of synthetic samples. We also evaluate the performance of compared models on our test protocol. Since there is no ground truth in real cases, we evaluate the results with IS \cite{salimans2016improved}, which is often used to measure the sample quality of GAN-based methods. Besides, we follow the evaluation using style loss \cite{gatys2016neural} in \cite{zeng2022sketchedit}, and compute the distance between feature maps from \texttt{relu\_1} and \texttt{relu\_2} in \cite{vgg}, which are denoted as SL\texttt{\_1} and SL\texttt{\_2}. Moreover, we conduct a user study to subjectively evaluate the compared methods in our supplementary materials.


\noindent\textbf{State-of-the-Art Competitors.} We compare our proposed method with both state-of-the-art image inpainting and image editing methods, including DeepFill-v2 \cite{yu2018free}, DeepPS \cite{yang2020deep}, LaMa \cite{suvorov2022resolution}, SketchEdit \cite{zeng2022sketchedit}, and ZITS \cite{dong2022incremental}. We re-train DeepFill-v2 \cite{yu2018free}, LaMa \cite{suvorov2022resolution} and ZITS \cite{dong2022incremental} with the same iterations as ours for fair comparisons. We use official model weights of DeepPS \cite{yang2020deep} and SketchEdit \cite{zeng2022sketchedit}. Note that the mask estimator of \cite{zeng2022sketchedit} could sometimes predict incorrect masks as we observe. For fair comparisons, we provide both quantitative results using the estimated masks and the same masks as ours.

\subsection{Performance Evaluation}
\noindent\textbf{Quantitative Comparison.} Table~\ref{table1} reports the quantitative results with the synthetic samples on CelebA-HQ \cite{karras2018progressive}, ImageNet \cite{russakovsky2015imagenet}, and Places \cite{zhou2017places}. Table~\ref{table2} reports the quantitative results on our test protocol using qualitative metrics. Our proposed method outperforms other competitors in all quantitative metrics.

\begin{figure*}[t]
    \centering
    \includegraphics[width=1.0\textwidth]{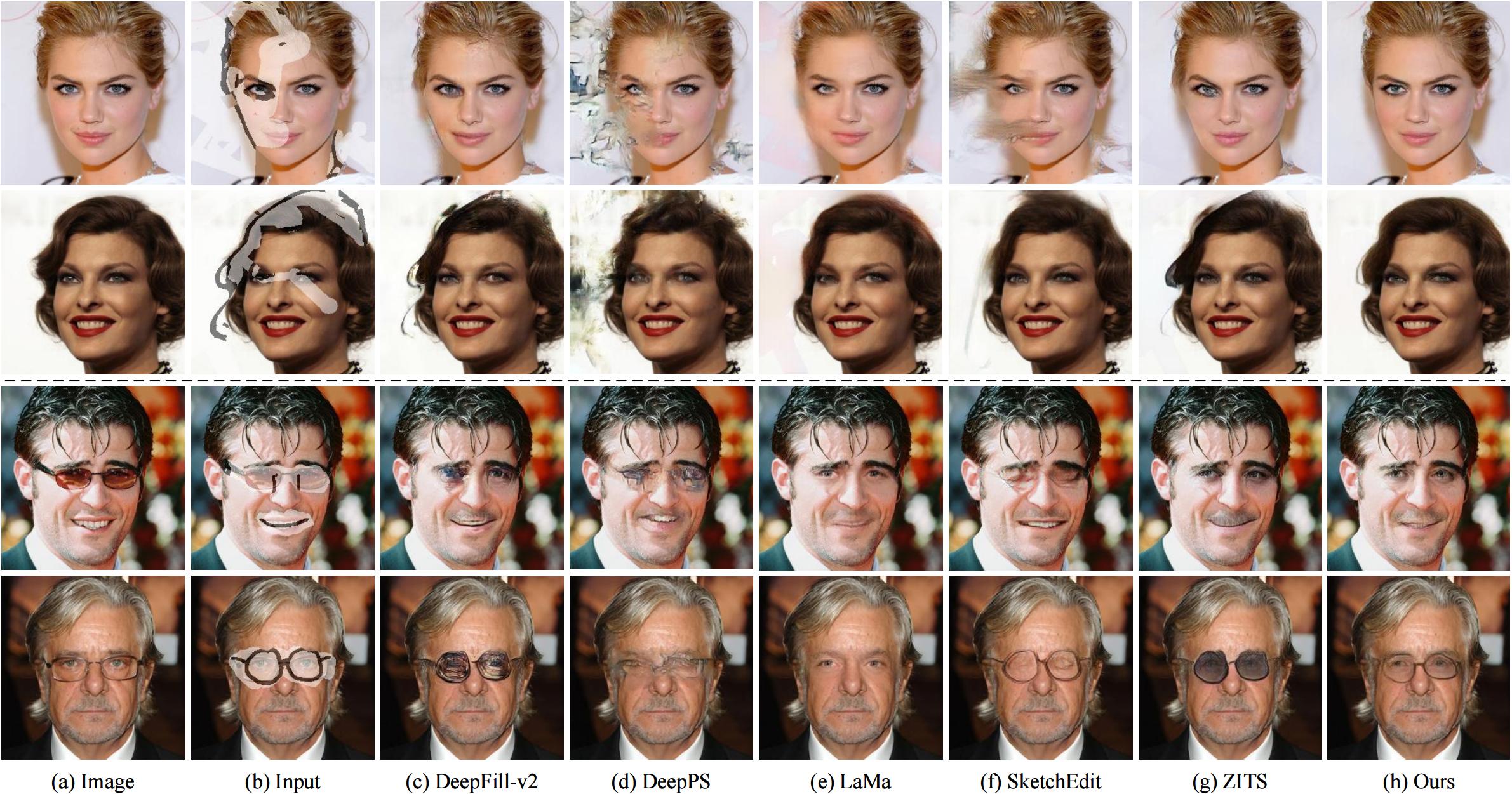} 
    \caption{Qualitative results of face restoration with synthetic sketches (top two rows) and face editing with user-drawn ones (bottom two rows), compared to (c) DeepFill-v2 \cite{yu2018free}, (d) DeepPS \cite{yang2020deep}, (e) LaMa \cite{suvorov2022resolution}, (f) SketchEdit \cite{zeng2022sketchedit}, and (g) ZITS \cite{dong2022incremental}. (b) represents the visualization of masked input with sketch. \textbf{Zoom in for best view.}}
    \label{fig3}
    \vspace{3em}
    \end{figure*}

\begin{figure*}[t!]
\centering
\includegraphics[width=1.0\textwidth]{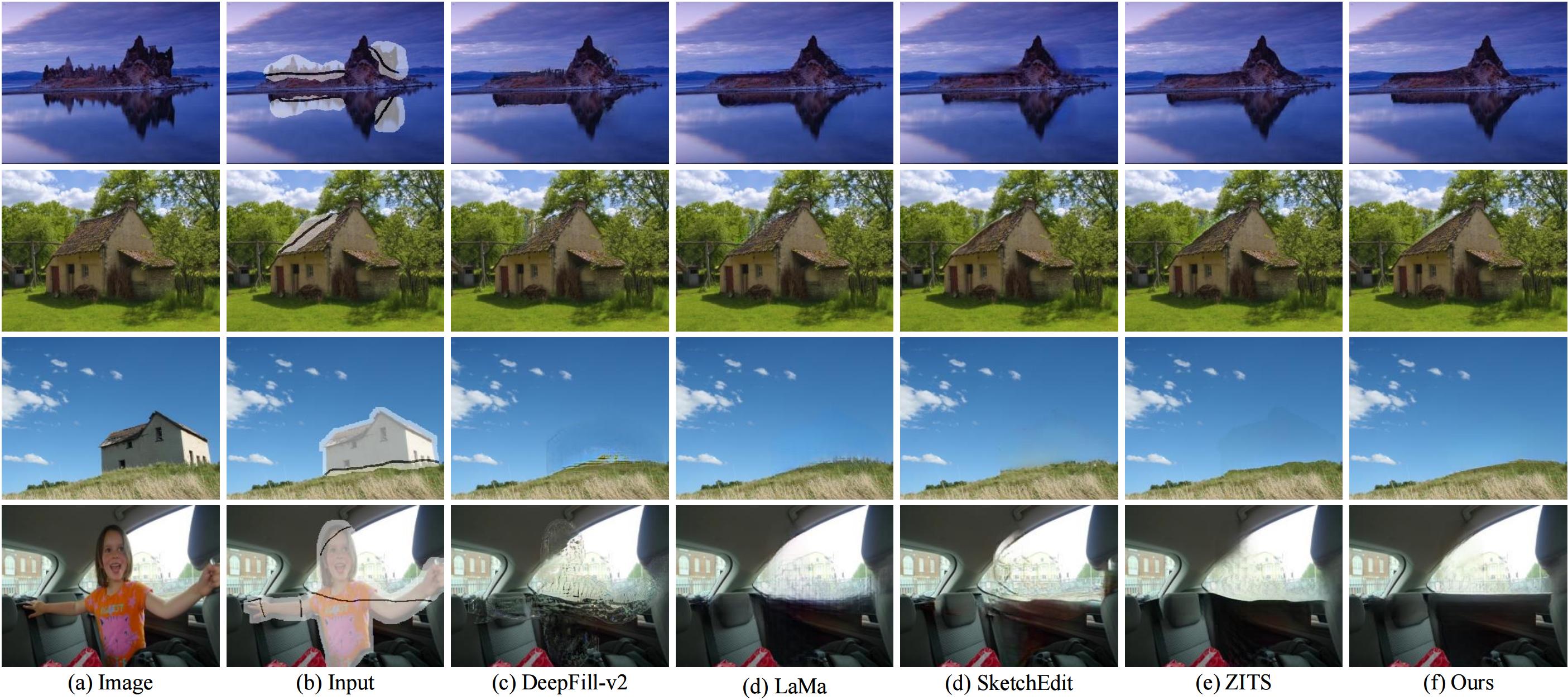} 
\caption{Qualitative results of scene editing with user-drawn sketches, compared to (c) DeepFill-v2 \cite{yu2018free}, (d) LaMa \cite{suvorov2022resolution}, (e) SketchEdit \cite{zeng2022sketchedit}, and (f) ZITS \cite{dong2022incremental}. (b) represents the visualization of masked input with sketch. \textbf{Zoom in for best view.}}
\label{fig4}
\end{figure*}

\noindent\textbf{Qualitative Comparison.} We evaluate the performance of the compared methods with both simulated samples and user interactions. Fig.~\ref{fig3} and~\ref{fig4} show the qualitative comparison among all methods of face restoration, face manipulation and scene editing. Given synthetic sketches or user-drawn ones, DeepFill-v2 \cite{yu2018free}, SketchEdit \cite{zeng2022sketchedit} and ZITS \cite{dong2022incremental} tend to produce unnatural results without the exact edge information. DeepPS \cite{yang2020deep} fails to accept the user-drawn sketches and causes severe artifacts. LaMa \cite{suvorov2022resolution} fails to recover the structures in the mask region without sketch guidance. Our method preserves user intuition with a proper magnitude of sketch refinement and produces plausible results against other competitors.

\subsection{Ablation Study}
\label{AblationStudy}

\noindent\textbf{Effect of Sketch Refinement.} Fig.~\ref{alignment} shows qualitative results for ablation studies of sketch refinement. We set the input sketch as an all-zero map to ablate the effect without sketch guidance. It could be seen that results appear to be blur and structurally disordered without sketch guidance. Using unrefined sketch leads to more details in the inpainting results, but still causes artifacts without sketch refinement. Guided with refined sketches, our method produces more plausible results with explicit outlines.

\begin{figure}[t!]
\centering
\includegraphics[width=0.825\linewidth]{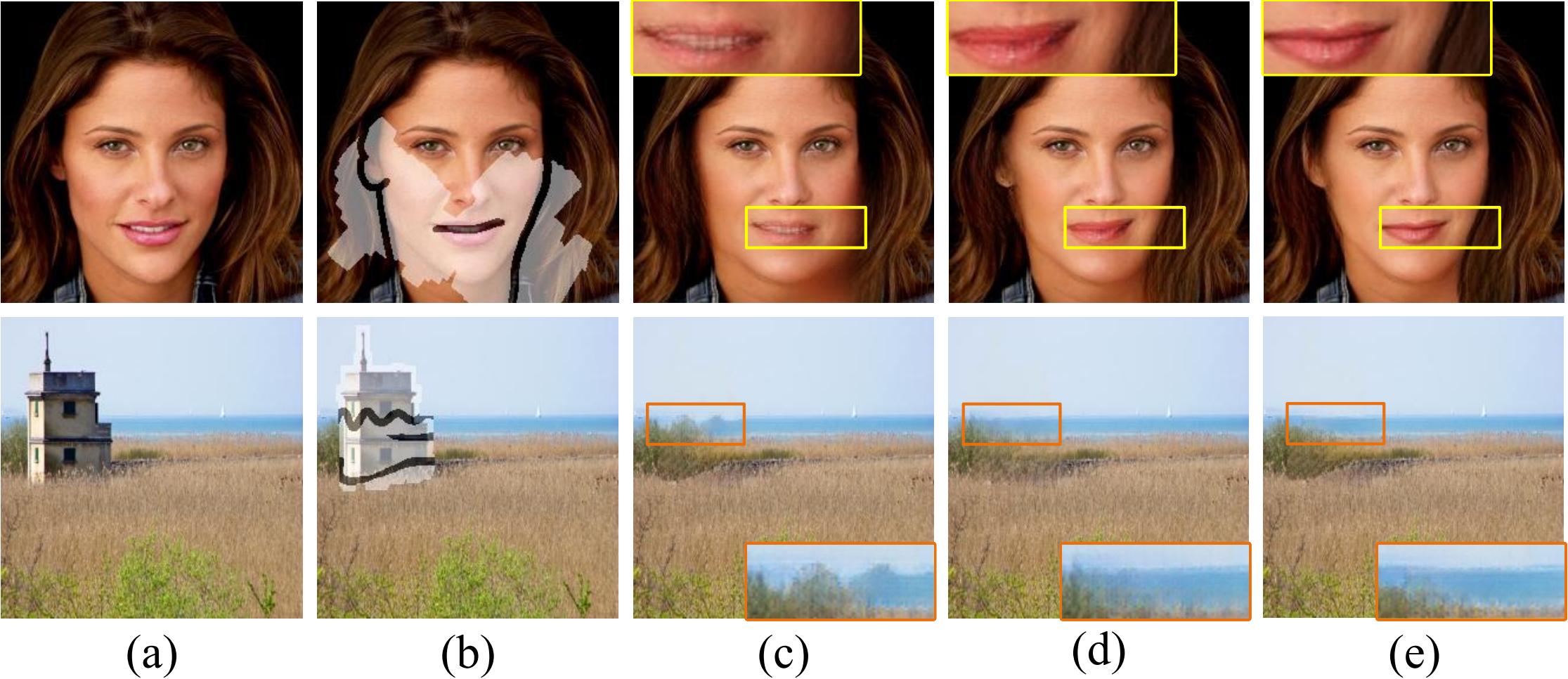} 
\caption{Qualitative results for the ablation studies of SRN, corresponding to Table~\ref{table3}. (a) Image, (b) input, (c) using RM, (d) using RM and EM, (e) using RM, EM, and $\mathcal{L}_{cc}$. (b) represents the visualization of masked input with sketch. \textbf{Zoom in for best view.}}
\label{fig5}
\end{figure}

\begin{figure}[t!]
\centering
\includegraphics[width=1.0\linewidth]{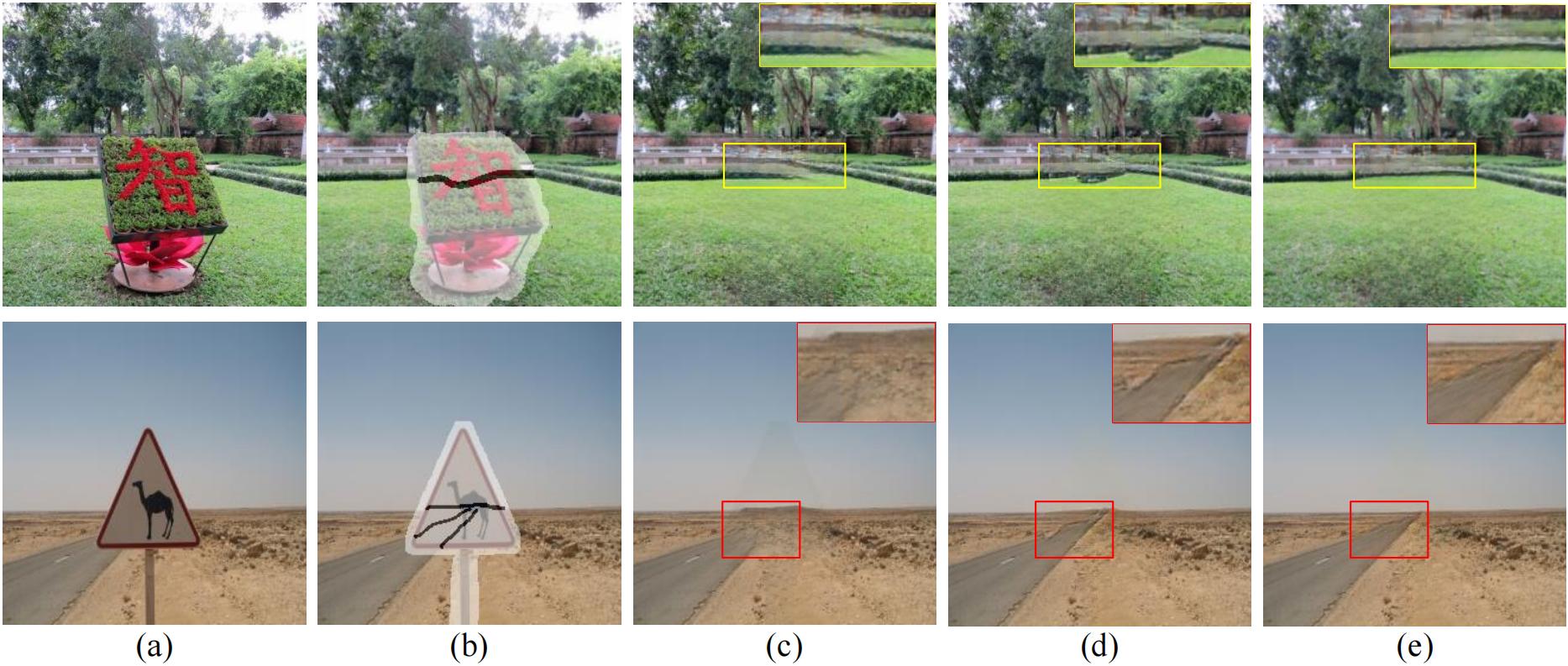} 
\caption{Qualitative results for ablation studies of sketch refinement. (a) Image, (b) input, (c) result without sketch, (d) result using unrefined sketch, (e) result using refined sketch. (b) represents the visualization of masked input with sketch. \textbf{Zoom in for best view.}}
\vspace{-1em}
\label{alignment}
\end{figure}

\noindent\textbf{Architectural Design of SRN.} In SRN, RM and EM are vital to handle the \emph{misalignment} and \emph{incoherence} correspondingly. Fig.~\ref{fig5} and Table~\ref{table3} show corresponding results. By comparing first two rows of Table~\ref{table3}, (c) and (d) in Fig.~\ref{fig5}, abandoning EM in SRN causes incoherence in the produced results and a performance drop in all quantitative metrics.

\noindent\textbf{Cross-Correlation Loss.} Cross-correlation loss is crucial to obtain guaranteed performance of SRN. By comparing last two rows in Table~\ref{table3}, training SRN without $\mathcal{L}_{cc}$ leads to obvious performance drop in all metrics. Comparing (d) and (e) in Fig.~\ref{fig5}, training with only $\mathcal{L}_{\ell_1}$ would be insufficient to handle the region-level \emph{misalignment} and \emph{incoherence} of sketches, and cause artifacts in qualitative results. Also, it is worth mentioning that our cross-correlation loss does not bring any extra computation cost during inference, meanwhile ensuring the superior performance of SRN. 

\begin{table}[t!]
    \centering
    \setlength{\tabcolsep}{0.7mm}
    \small
    \begin{tabular}{cccccc}
     \toprule
     \multicolumn{3}{c}{Components of SRN}&\multicolumn{3}{c}{CelebA-HQ}\\
     \quad $\mathcal{L}_{cc}$\quad&\quad EM \quad&\quad RM\quad&\quad PSNR$^\uparrow$\quad&\quad SSIM$^\uparrow$\quad&\quad FID$^\downarrow$\quad \\
     \cmidrule(r){1-3}\cmidrule(r){4-6}
     &&\quad$\checkmark$&\ \ 23.60\quad &\ \ 0.828\quad &\ \ 20.81\quad\\
      &\quad$\checkmark$&\quad$\checkmark$&\ \ 23.93\quad &\ \ 0.831\quad &\ \ 16.66\quad\\
     \quad$\checkmark$&\quad$\checkmark$&\quad$\checkmark$&\ \ \textcolor{red}{24.99}\quad &\ \ \textcolor{red}{0.877}\quad &\ \ \textcolor{red}{5.537}\quad \\
     \bottomrule
    \end{tabular}
    \caption{Quantitative results for ablation studies of $\mathcal{L}_{cc}$ and EM with synthetic samples on CelebA-HQ \cite{karras2018progressive}, corresponding to Fig.~\ref{fig5}. Here, the \textcolor{red}{best} results are in red.}
    \label{table3}
\end{table}

\begin{figure}[t!]
\centering
\includegraphics[width=1.0\linewidth]{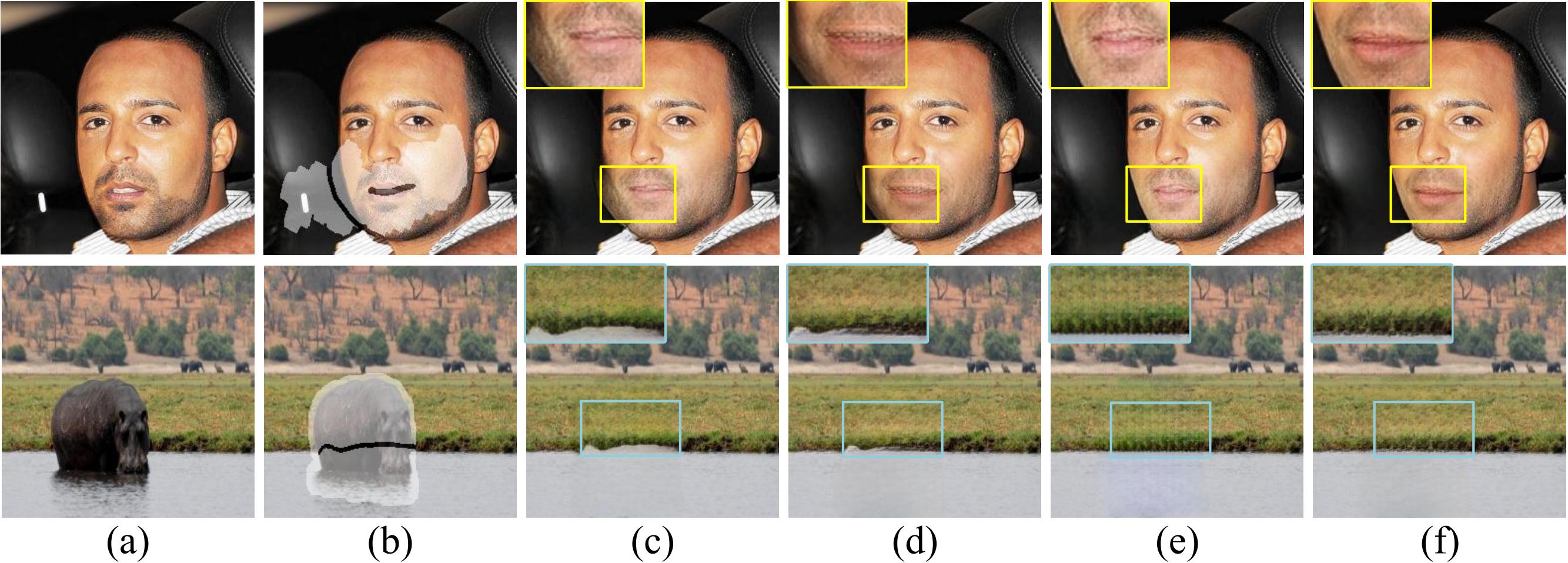} 
\caption{Qualitative results for the ablation studies of SketchRefiner, corresponding to Table~\ref{table4}. (a) Image, (b) input, (c) using neither SRN nor SFA, (d) using SFA, (e) using SRN, (f) using both SRN and SFA. (b) represents the visualization of masked input with sketch. \textbf{Zoom in for best view.}}
\label{fig6}
\end{figure}

\noindent\textbf{SRN.} SRN plays a vital role in bridging the gap between sketch-like inputs and image inpainting methods. Fig.~\ref{fig6} and Table~\ref{table4} show corresponding results. In Fig.~\ref{fig6}, we could see that calibrating sketches with SRN allows SIN to diminish the artifacts due to \emph{misalignment} and \emph{incoherence} of the input sketches. Comparing the second and the fourth row of Table~\ref{table4}, it could be seen that SRN brings a significant performance gain in all quantitative metrics.

\begin{table}[t!]
    \centering
    \setlength{\tabcolsep}{0.7mm}
    \small
    \begin{tabular}{ccccc}
     \toprule
     \multicolumn{2}{c}{Components of SketchRefiner}&\multicolumn{3}{c}{ImageNet}\\
     \quad SRN \quad&\quad SFA \quad&\quad PSNR$^\uparrow$\quad&\quad SSIM$^\uparrow$\quad&\quad FID$^\downarrow$\quad \\
     \cmidrule(r){1-2}\cmidrule(r){3-5}
     &&\ \ 20.46\quad &\ \ 0.745\quad &\ \ 22.34\quad\\
     &\quad$\checkmark$\quad&\ \ 21.03\quad &\ \ 0.749\quad &\ \ 16.03\quad\\
     \quad$\checkmark$\quad&&\ \ 21.29\quad &\ \ 0.750\quad &\ \ 15.37\quad\\
     \quad$\checkmark$\quad&\quad$\checkmark$\quad&\ \ \textcolor{red}{22.16}\quad &\ \ \textcolor{red}{0.769}\quad &\ \ \textcolor{red}{9.531}\quad\\
     \bottomrule
    \end{tabular}
    \caption{Quantitative comparison for ablation studies of SRN and SFA with synthetic samples on ImageNet \cite{russakovsky2015imagenet}, corresponding to Fig.~\ref{fig6}. Here, the \textcolor{red}{best} results are in red.}
  \label{table4}
\end{table}
\noindent\textbf{SFA Block.} SFA block is essential of aggregating the extracted features as modulation in SIN. Comparing (e) and (f) in Fig.~\ref{fig6}, last two rows in Table~\ref{table4}, using the original design in \cite{suvorov2022resolution} simply causes high frequency disorder, which leads to performance drop in all metrics quantitatively.

\section{Conclusion} In this paper, we re-investigate the challenges of sketch-based interactive image inpainting. We attempt to bridge the gap between sketch-like inputs and current image inpainting models by proposing a two-stage system, called SketchRefiner. SketchRefiner is able to restore complex structures in the corrupted regions and reveals great potential in real-world applications. Moreover, we establish an algorithm for sketch simulation and propose a sketch-based test protocol for real-world applications. Experiments on synthetic and user-provided samples demonstrate the superiority of our SketchRefiner both qualitatively and quantitatively.

SketchRefiner is still limited and tends to over-refine sketches with meticulous structures. Our supplementary materials provide more discussions on failure cases and working boundaries of our SketchRefiner. More future work is anticipated to balance the requirements of sketch controlling ability and tolerance of sketch randomness.

\clearpage

{\small
\bibliographystyle{ieee_fullname}
\bibliography{main_paper}
}

\clearpage



\setcounter{section}{0}
\renewcommand\thesection{\Alph{section}}
\section{Overview}
We manage to formulate our supplementary materials as follows. In Sec.~\ref{ImplementationDetails}, we introduce the implementation details of our Sketch Simulation Algorithm (SSA), the loss functions, the training strategies of retrained methods and ours, and our proposed test protocol. In Sec.~\ref{NetworkArchitecture}, we provide detailed architectural designs of each component in SketchRefiner. In Sec.~\ref{FailureCases}, we discuss the limitations and working boundaries of SketchRefiner. In Sec.~\ref{ExtendedExperiments}, we report our conducted user study for perceptual evaluation, the model scalability to other inpainting methods, the ablation studies for our Partial Sketch Encoder (PSE), and the ablation studies for Sketch Feature Aggregation (SFA) blocks. In Sec.~\ref{MoreResults}, we show the visualization of the refined sketches, and more results of the proposed technique under real-world applications in the proposed testing benchmark. In Sec.~\ref{InteractiveDemo}, we introduce our designed interactive demo for end users to experience the process of sketch refinement and interactive image inpainting.

\section{Implementation Details}
\label{ImplementationDetails}

\subsection{Sketch Simulation Algorithm}
The proposed SSA aims to simulate \emph{misalignment}, \emph{incoherence}, and \emph{abstraction} of sketches. In the simulation of \emph{misalignment}, it essentially projects pixels in the convolution kernel to other ones fitting a Gaussian distribution. Referring to Algorithm 1 in our main paper, we set $k=11$ as the kernel size of the Gaussian Filter (GF). We use $\psi$ to manipulate the magnitude of warping. For the selection of the deformation magnitude $\psi$, we consider the spatial variation of the recomposed sketch $\textbf{S}$ according to the different demands in real-world cases, and manage to adapt to these scenarios by varying the values of $\psi$ during training. For instance, when handling cases that tend to manipulate the image contents such as face manipulation or scene editing, $\textbf{S}$ has obvious variations compared to the edge map $\textbf{E}$ of the ground truth image, thus leading to a higher value of $\psi$ and a more intensified magnitude of deformation. When handling cases that tend to recover missing textures such as background restoration or object removal, we expect $\textbf{S}$ to be as close to $\textbf{E}$ as possible, which causes less variation and the value of $\psi$ to its smallest extent. Particularly, such selection of $\psi$ also considers the case that it is tough for untrained users to provide the pixel-wise aligned outlines, and simulate the mild disruption in the pixel-wise preciseness of $\textbf{S}$. Practically, we sample $\psi$ from a uniform distribution $U[\psi_{min}, \psi_{max}]$ within every training batch, and set $\psi_{min}=0.01, \psi_{max}=0.8$ for best performance.

\subsection{Loss Functions}
We mainly illustrate the loss-weighing hyperparameters of the mentioned loss functions in this paragraph. As for the loss functions of the Registration Module (RM) and Enhancement Module (EM) in Sketch Refinement Network (SRN), we write $\mathcal{L}_{RM}=\mathcal{L}_{\ell_1}+\lambda_1\mathcal{L}_{cc}$ and $\mathcal{L}_{EM}=\mathcal{L}_{\ell_1}+\lambda_2\mathcal{L}_{cc}$, where $\lambda_1=0.4$ and $\lambda_2=0.9$ for best performance. As for the loss function of Sketch-modulated Inpainting Network (SIN), termed $\mathcal{L}_{SIN} = \lambda_{\ell_1}\mathcal{L}_{\ell_1} + \lambda_{adv}\mathcal{L}_{adv} + \lambda_{fm}\mathcal{L}_{fm} + \lambda_{HRF}\mathcal{L}_{HRF}$, we set $\lambda_{\ell_1} = 10, \lambda_{adv} = 10, \lambda_{fm} = 100, \lambda_{HRF} = 30$ to obtain its best performance.

\subsection{Training Strategies}

As for retraining of other compared methods, including DeepFill-v2 \cite{yu2018free}, LaMa \cite{suvorov2022resolution}, and ZITS \cite{dong2022incremental}, we follow the reported strategy in their paper to achieve their best performance. We only preserve the part of the inpainting network of ZITS for evaluation. Note that the reported optimization step of Deep-Fillv2 is 300k steps and the one of ZITS is 800k steps on Places \cite{zhou2017places}. Thus, retraining with 800k steps should be sufficient. DeepFill-v2 \cite{yu2018free} uses the edge map detected by HED \cite{xie2015holistically} for training, and ZITS \cite{dong2022incremental} uses canny edge maps together with detected lines, which differs greatly from our simulated ones. For fair comparisons, we use edge maps detected from the ground truth images by \cite{he2020bdcn} for their training. LaMa \cite{suvorov2022resolution} trains their models for 1M iterations. Since we do not focus on the large mask setting, we retrain it for 800k iterations following the mask setting of \cite{yu2018free}, which has smaller mask regions and easier for the model to get converged. 

As for the training of our SketchRefiner, we train the SRN with an NVIDIA 2080Ti GPU, and the SIN with two NVIDIA GeForce 3090 GPUs. As for the training details of the SRN, we use the Adam \cite{kingma2014adam} optimizer with a learning rate of 1e-4. We train SRN for 500k steps on ImageNet \cite{russakovsky2015imagenet} with free-form mask \cite{yu2018free} and the simulated sketch using our SSA. The batch size of SRN is set as 10. It is worth mentioning that our trained SRN reveals remarkable generalization to unseen datasets, e.g., CelebA-HQ \cite{karras2018progressive} and Places \cite{zhou2017places} without retraining. As for the training details of SIN, we set the learning rate of the generator as 3e-4, and the one of the discriminator as 1e-4 using Adam \cite{kingma2014adam} optimizer. The learning rates are reduced to half in the middle of the training process. We train SIN for 800k steps on ImageNet \cite{russakovsky2015imagenet} and Places \cite{zhou2017places}, and for 400k steps on CelebA-HQ \cite{karras2018progressive}. The batch size of SIN is set as 25. During training, all images, masks, sketches, and edges are resized to $256\times256$ resolution. Note that the sketches and edges are all binarized with 0-1 binary thresholding. And our codebase is implemented based on PyTorch and will be open-sourced.

\begin{figure}[t!]
\centering
\includegraphics[width=1.0\linewidth]{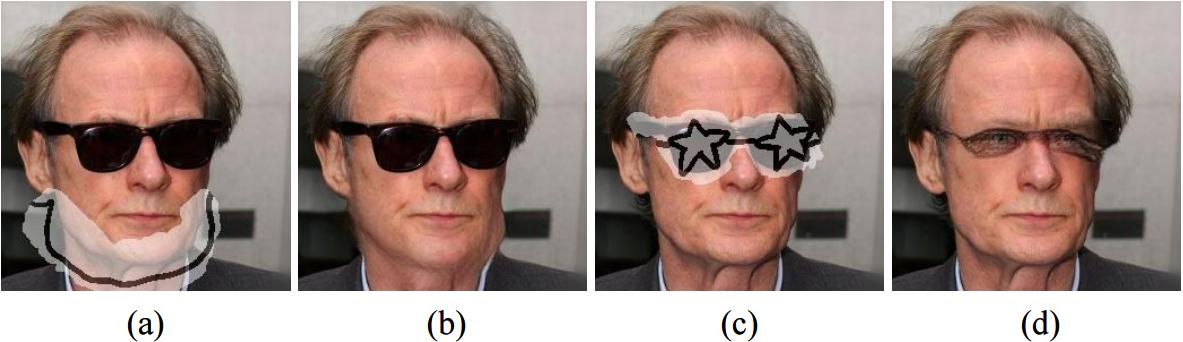} 
\caption{Failure cases of our SketchRefiner. (From left to right) (a) Input, (b) result, (c) input, (d) result. Note that (a) and (c) represent the visualization of masked input with sketch. \textbf{Zoom in for best view.}}
\label{fig:failure_cases}
\end{figure}

\begin{table}[t!]
    \centering
    \setlength{\tabcolsep}{1.0mm}
    \caption{Average top-1 percentage of the conducted user study on editing cases of both face images and scene-level images, compared among DeepPS \cite{yang2020deep}, DeepFill-v2 \cite{yu2018free}, LaMa \cite{suvorov2022resolution}, SketchEdit \cite{zeng2022sketchedit} and ZITS \cite{dong2022incremental}. Here, the \textcolor{red}{best} and \textcolor{blue}{second best} results are in red and blue. ``-'' stands for unavailable results because the corresponding methods did not perform experiments on that dataset.}
    \scalebox{0.9}{
    \begin{tabular}{lccc}
     \toprule
     & \multicolumn{3}{c}{Test Protocol} \\
     \quad Methods \quad & \quad Face \quad & \quad Scene \quad & \quad Total \quad \\
	 \midrule     
     DeepPS \cite{yang2020deep} & \quad 10.61\% \quad & \quad - \quad & \quad - \quad \\
     DeepFill-v2 \cite{yu2018free} & \quad 10.61\% \quad & \quad 11.36\% \quad & \quad 11.56\% \quad \\
     LaMa \cite{suvorov2022resolution} & \quad 12.12\% \quad & \quad 15.91\% \quad & \quad 14.97\% \quad \\
     SketchEdit \cite{zeng2022sketchedit} & \quad 16.67\% \quad & \quad 7.955\% \quad & \quad 12.24\% \quad \\
     ZITS \cite{dong2022incremental} & \quad \textcolor{blue}{18.18\%} \quad & \quad \textcolor{blue}{22.73\%} \quad & \quad \textcolor{blue}{21.77\%} \quad \\
     \textbf{Ours} & \quad \textcolor{red}{31.81\%} \quad & \quad \textcolor{red}{42.05\%} \quad & \quad \textcolor{red}{39.46\%} \quad \\
     \bottomrule
    \end{tabular}
    }
    \label{table_user_study}
\end{table}

\subsection{Test Protocol}
\label{TestProtocol}

Previous inpainting methods tend to evaluate the performance of models in a self-supervised manner, which uses synthetic masks to corrupt the ground truth and require the model to recover it. Such simulation is similar to the cases of removing unexpected foregrounds or occlusions in real-world applications. However, it actually occurs that we cannot acquire the ground truth during a real-time interface. This indicates the absence of testing benchmarks evaluating models with user-provided samples. To this end, we provide our solution by collecting a sketch-based test protocol. The collected test protocol consists of 100 samples in total, with 30 face images and 70 scene-level images, respectively. It focuses on both editing cases of scene-level images and face images, and is applicated to sketch-based real-world editing cases, e.g., scene editing, face manipulation, object removal, and background restoration. Both mask and sketch of a corresponding image are manually annotated using the application SketchBook on an iPad with an Apple Pencil. Fig.~\ref{RealWorldSamples} shows a number of samples of the collected test protocol.

\begin{figure*}[t!]
    \centering
    \includegraphics[width=1.0\linewidth]{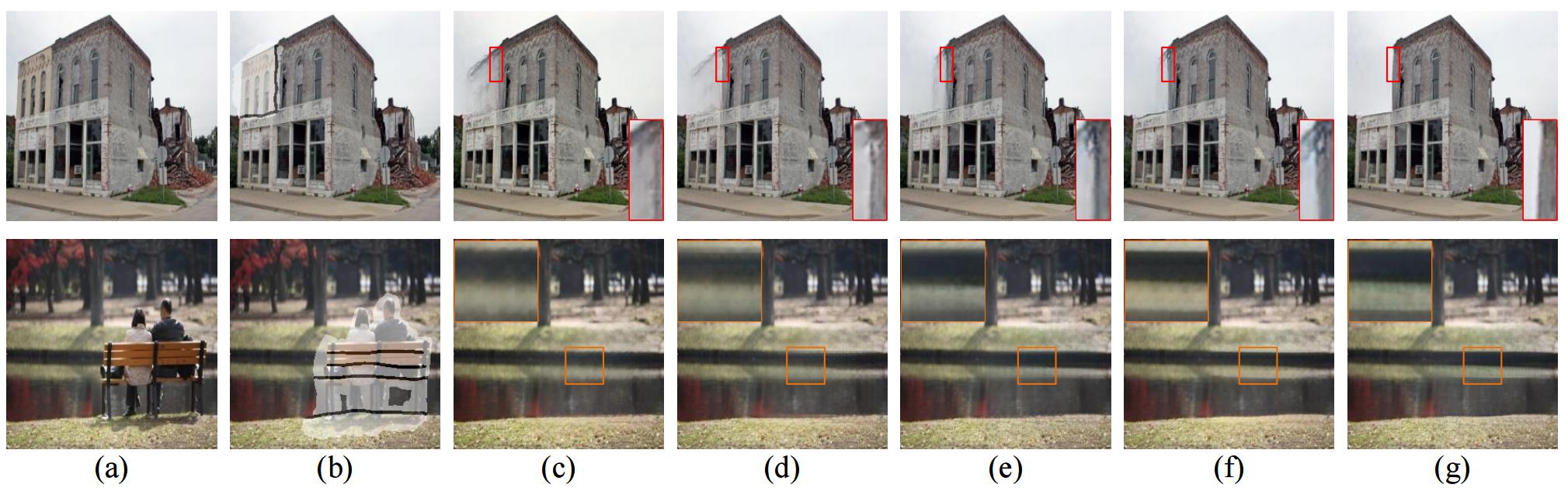} 
    \caption{Qualitative results of scene editing for the ablation studies of coarse-to-fine guiding effect in PSE. (From left to right) (a) Image, (b) input, (c) \emph{concatenate} sketch as an additional channel instead of using PSE, (d) using $\mathcal{F}_{n-3}$, (e) using $\mathcal{F}_{n-3}$ and $\mathcal{F}_{n-2}$, (f) using $\mathcal{F}_{n-3}$, $\mathcal{F}_{n-2}$, and $\mathcal{F}_{n-1}$, and (g) using all extracted features $\mathcal{F}$ in PSE. \textbf{Zoom in for best view.}}
    \label{fig:pse}    
\end{figure*}

\section{Network Architecture}
\label{NetworkArchitecture}
We provide the detailed network architecture of both SRN and SIN. Tab.~\ref{network_rm} and Tab.~\ref{network_em} show the detailed architecture of the Registration Module (RM) and Enhancement Module (EM) in SRN. Tab.~\ref{network_pse} and Tab.~\ref{network_trm} show the detailed architecture of the Partial Sketch Encoder (PSE) and Texture Restoration Module (TRM) in SIN. 

\section{Failure Cases and Discussions}
\label{FailureCases}
Fig.~\ref{fig:failure_cases} shows some failure cases of SketchRefiner. As is shown in (b) of Fig.~\ref{fig:failure_cases}, SketchRefiner fails to calibrate misaligned sketches with large mask regions, thus causing blur and artifacts in the generated result. One possible solution for this problem might be enlarging the receptive field of SRN. (d) reveals that SketchRefiner sometimes over-refines sketches with meticulous structures, e.g., the star-shaped sunglasses. This also indicates the diversities and challenges of sketch refinement among real cases. Besides, SketchRefiner only accepts monochrome sketches as inputs currently, since the training data and our PSE are designed for binary sketches. This might be limited to some further image editing scenarios, e.g., image manipulation with colored strokes. 

\section{Extended Experiments}
\label{ExtendedExperiments}

\noindent\textbf{User Study.} In addition to the qualitative metrics introduced in our main paper, we also conduct a user study to evaluate the compared methods perceptually. In this study, we recruit 11 unique volunteers as participants. We first show each participant the original images, corresponding masks, user-drawn sketches, and results produced by all compared methods anonymously. In every editing case, each participant is requested to pick the result that is most visually plausible. We randomly sample 28 editing cases in our test protocol, and show the sampled cases to all participants. The samples consist of 16 cases of scene-level images and 12 cases of face images. Afterward, we calculate the average top-1 percentage for each method as the final rate. Tab.~\ref{table_user_study} reports the results of the conducted user study. Users overwhelmingly prefer the results produced by our SketchRefiner rather than other competitors in all cases.

\begin{table}[t!]
    
    \centering
    \caption{Quantitative results of the model scalability upon other state-of-the-art methods on CelebA-HQ \cite{karras2018progressive}, including DeepFill-v2 \cite{yu2018free}, SketchEdit \cite{zeng2022sketchedit}, and ZITS \cite{dong2022incremental}. Experiments using the same method represent a comparison group. Here, the \textcolor{red}{best} result in each group is in red.} 
    \small
    \begin{tabular}{ccccc}
     \toprule
     \multicolumn{2}{c}{}&\multicolumn{3}{c}{CelebA-HQ}\\
     \quad Methods \quad&\quad SRN \quad&\quad PSNR$^\uparrow$\quad&\quad SSIM$^\uparrow$\quad&\quad FID$^\downarrow$\quad \\
     \cmidrule(r){1-2}\cmidrule(r){3-5}
     DeepFill-v2 \cite{yu2018free} &&\ \ 22.06\quad &\ \ 0.788\quad &\ \ 26.78\quad\\
     DeepFill-v2 \cite{yu2018free} &\quad$\checkmark$\quad&\ \ \textcolor{red}{23.61}\quad &\ \ \textcolor{red}{0.817}\quad &\ \ \textcolor{red}{24.72}\quad\\
     SketchEdit \cite{zeng2022sketchedit} &&\ \ 20.21\quad &\ \ 0.759\quad &\ \ 41.28\quad\\
     SketchEdit \cite{zeng2022sketchedit} &\quad$\checkmark$\quad&\ \ \textcolor{red}{22.62}\quad &\ \ \textcolor{red}{0.813}\quad &\ \ \textcolor{red}{20.15}\quad\\
     ZITS \cite{dong2022incremental} &&\ \ 22.79\quad &\ \ 0.812\quad &\ \ 18.79\quad\\
     ZITS \cite{dong2022incremental} &\quad$\checkmark$\quad&\ \ \textcolor{red}{23.60}\quad &\ \ \textcolor{red}{0.830}\quad &\ \ \textcolor{red}{17.63}\quad\\
     \bottomrule
    \end{tabular}
  \label{PlugAndPlayTable}
\end{table}

\begin{table}[t!]
    \centering
    \caption{Quantitative results for the ablation studies of coarse-to-fine guiding effect in PSE on CelebA-HQ \cite{karras2018progressive}. Here, the \textcolor{red}{best} result is in red. \emph{Concat.} represents that we follow the solution of edge-based methods \cite{Nazeri_2019_ICCV, yu2018free, cao2021learning} by concatenating sketches as an additional channel of the inpainting network. The ticked $\mathcal{F}_{i}$ represents the extracted features in $i$-th layer of PSE that we send into the inpainting network. $n$ denotes the network depth of PSE.}
    \setlength{\tabcolsep}{0.05mm}
    \small
    \scalebox{0.9}{\begin{tabular}{cccccccc}
     \toprule
     \multicolumn{5}{c}{Methods} & \multicolumn{3}{c}{CelebA-HQ}\\
     \quad $\mathcal{F}_{n-3}$ \quad&\quad $\mathcal{F}_{n-2}$ \quad&\quad $\mathcal{F}_{n-1}$ \quad&\quad $\mathcal{F}_{n}$ \quad&\quad \emph{Concat.} \quad&\quad PSNR$^\uparrow$\quad&\quad SSIM$^\uparrow$\quad&\quad FID$^\downarrow$\quad \\
     \cmidrule(r){1-5}\cmidrule(r){6-8}
     \quad \quad & \quad \quad &\quad \quad & \quad \quad &\quad $\checkmark$ \quad & \quad 19.58 \quad & \quad 0.681 \quad & \quad 25.97 \quad \\
     \quad \quad & \quad \quad &\quad \quad & \quad $\checkmark$ \quad & \quad \quad & \quad 20.76 \quad & \quad 0.709 \quad & \quad 17.11 \quad\\
     \quad \quad & \quad \quad &\quad $\checkmark$ \quad & \quad $\checkmark$ \quad & \quad \quad & \quad 21.23 \quad & \quad 0.731 \quad & \quad 13.83 \quad\\
     \quad \quad & \quad $\checkmark$ \quad &\quad $\checkmark$ \quad & \quad $\checkmark$ \quad &\quad \quad & \quad 21.83 \quad & \quad 0.757 \quad & \quad 11.48 \quad\\
     \quad $\checkmark$ \quad & \quad $\checkmark$ \quad &\quad $\checkmark$ \quad & \quad $\checkmark$ \quad & \quad \quad & \quad \textcolor{red}{22.16} \quad & \quad \textcolor{red}{0.769} \quad & \quad \textcolor{red}{9.531} \quad\\
     \bottomrule
    \end{tabular}}
  \label{table:pse}
\end{table}

\noindent\textbf{Model Scalability.} Our proposed SketchRefiner is highly interpretable and scalable to other methods. We ablate its model scalability by implementing SRN upon other state-of-the-art methods. These methods include DeepFill-v2 \cite{yu2018free}, ZITS \cite{dong2022incremental}, and SketchEdit \cite{zeng2022sketchedit}. We first send the sketch-like inputs into our pre-trained SRN for sketch refinement. Then, we use the calibrated sketches during the inference of other methods. Tab.~\ref{PlugAndPlayTable} reports the quantitative results of our experiments on CelebA-HQ \cite{karras2018progressive}. Experiments demonstrate that using SRN in these methods contributes greatly to their performance in quantitative results due to the elimination of free-form randomness of sketches. Also, the SRN is capable of being a plug-and-play tool for other sketch-based image inpainting methods.

\begin{table}[t!]
    \centering
    \caption{Quantitative results for the ablation studies of the architectural design of SFA blocks on CelebA-HQ \cite{karras2018progressive}. \emph{Adding} indicates not using any design for aggregation by directly adding the sketch features to the image features. Here, the \textcolor{red}{best} result is in red.}
    \setlength{\tabcolsep}{1.2mm}
    \small
    \scalebox{1.0}{
    \begin{tabular}{cccc}
     \toprule
     & \multicolumn{3}{c}{CelebA-HQ} \\
     \quad Methods \quad&\quad PSNR$^\uparrow$\quad&\quad SSIM$^\uparrow$\quad&\quad FID$^\downarrow$\quad \\
     \midrule
     \quad \emph{Adding} \quad & \quad 20.46 \quad & \quad 0.745 \quad & \quad 22.34 \quad \\
     \quad using SPADE \cite{park2019semantic} \quad & \quad 20.59 \quad & \quad 0.761 \quad & \quad 20.97 \quad \\
     \quad using SFA \quad & \quad \textcolor{red}{22.16} \quad & \quad \textcolor{red}{0.769} \quad & \quad \textcolor{red}{9.531} \quad\\
     \bottomrule
    \end{tabular}}
  \label{table:sfa}
\end{table}

\begin{figure}[t!]
\centering
\includegraphics[width=1.0\linewidth]{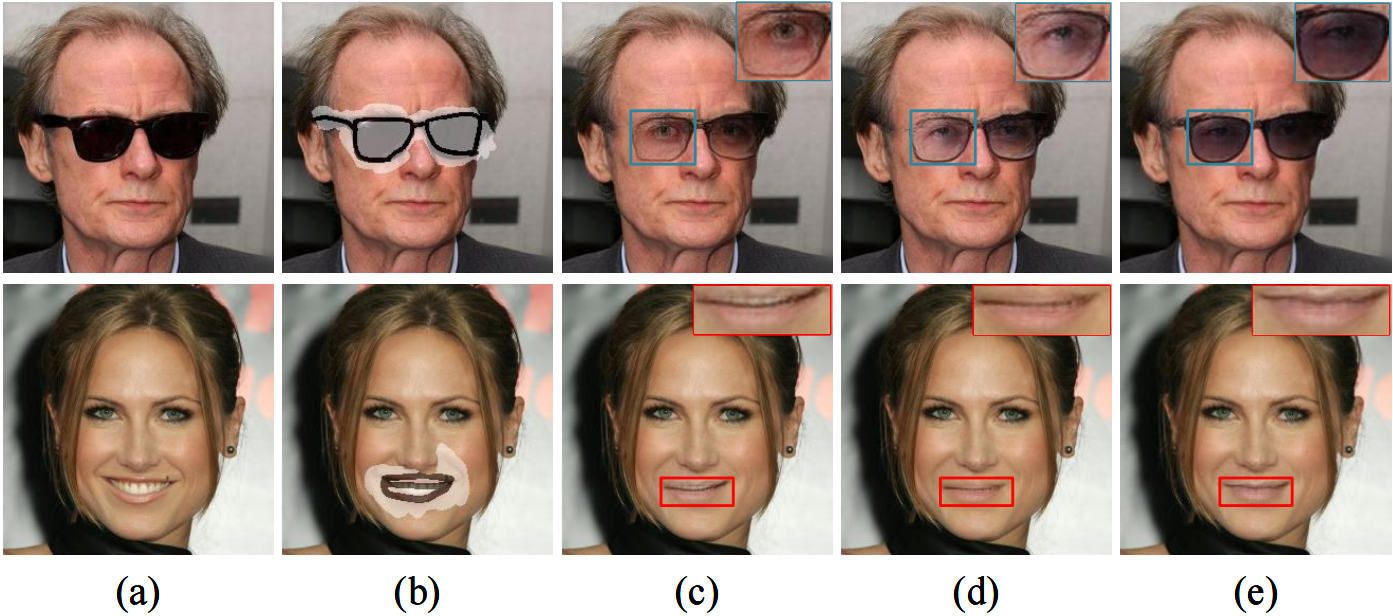} 
\caption{Qualitative results for the ablation studies of the architectural design of SFA. (From left to right) (a) Image, (b) input, (c) aggregating by \emph{adding}, (d) using SPADE \cite{park2019semantic}, and (e) using our SFA. \textbf{Zoom in for best view.}}
\label{fig:sfa}
\end{figure}

\noindent\textbf{Coarse-to-Fine Guiding Effect of PSE.} We ablate the coarse-to-fine guiding effect of PSE. We also investigate the art of injecting the sketch condition, by retraining our method following the pixel-level solution of previous methods \cite{Nazeri_2019_ICCV, yu2018free, cao2021learning}, concatenating the sketch as an additional channel of the inpainting network. Tab.~\ref{table:pse} shows the quantitative results for the ablation studies, and Fig.~\ref{fig:pse} shows the qualitative results. It could be seen that the previous pixel-level solution \cite{Nazeri_2019_ICCV, yu2018free, cao2021learning} fails to introduce the sketch condition and produce unguided results with artifacts, which is caused by the feature diminishing problem of concatenating \cite{liu2021deflocnet}. By increasing the number of features injected into the inpainting network, we could see a gradually improved performance and more explicit edges and textures in the generated content. It demonstrates that the coarse-to-fine guiding features in different layers of PSE provide multidimensional information in different granularity, and the effectiveness of PSE handling sketches in the latent space.

\noindent\textbf{Architectural Design of SFA.} We ablate the architectural design of our SFA, compared with the SPADE block \cite{park2019semantic}. Tab.~\ref{table:sfa} shows the quantitative results, and Fig.~\ref{fig:pse} shows the qualitative results for the ablation studies. If not using any specific design in the inpainting network to fuse the extracted features, the generated results show apparent color discrepancy and blur, which proves our argument that using the original layers in \cite{suvorov2022resolution} does not work well. Since SPADE \cite{park2019semantic} is designed for aggregating the semantic layout, using SPADE \cite{park2019semantic} in SketchRefiner leads to artifacts in the generated results due to the gap between different latent spaces. The solution of our SFA leads to optimal results.


\section{More Results}
\label{MoreResults}
We supply the visualization of the refined sketches and more results on our collected test protocol. Fig.~\ref{fig2} reports the visualization of the refined sketches and corresponding inpainting results.. Fig.~\ref{fig4} shows more results of face manipulation and scene editing upon the testing benchmark.

\section{Interactive Demo}
\label{InteractiveDemo}
We design an interactive demo for end users to experience the process of sketch refinement and interactive image inpainting based on Gradio \cite{abid2019gradio}. Fig.~\ref{fig:demo} shows the user interface of our interactive demo. Users need to first upload an image for manipulation. Then, they could custom their own mask and sketch by interactively drawing them on the left side. The inpainted results would be synchronously shown on the right side. By clicking the checkbox on the left side, the inpainted results with sketch refinement would be displayed. We provide a demo video of the interactive demo in our supplementary materials, and would release our interactive demo upon acceptance of our paper.

\begin{figure*}[t!]
    \centering
    \includegraphics[width=0.9\textwidth]{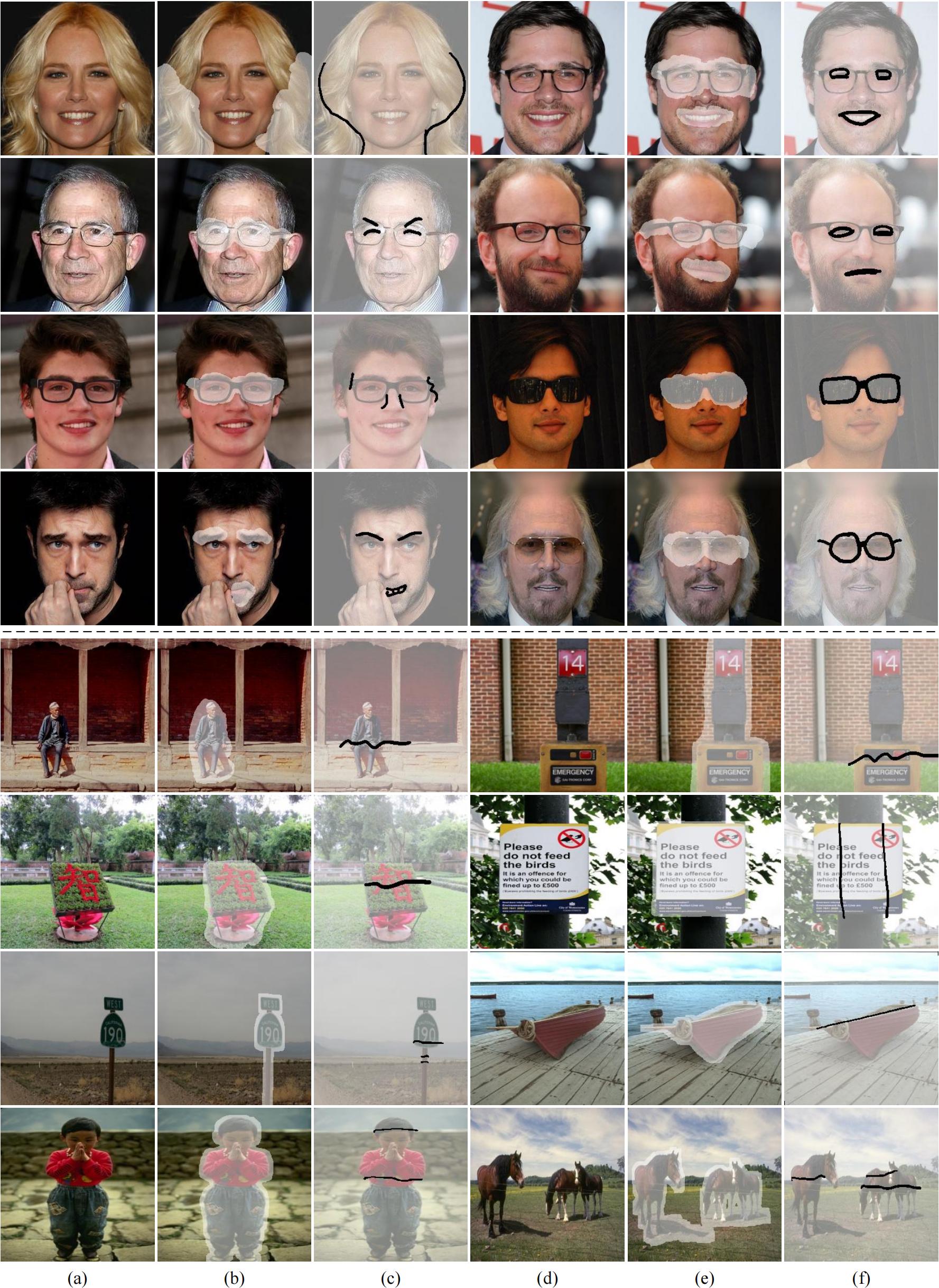} 
    \caption{Samples of the collected test protocol. (From left to right) (a) Image, (b) mask, (c) sketch, (d) image, (e) mask, (f) sketch.}
    \label{RealWorldSamples}
    \end{figure*}

\begin{figure*}[t!]
\centering
\includegraphics[width=1.0\linewidth]{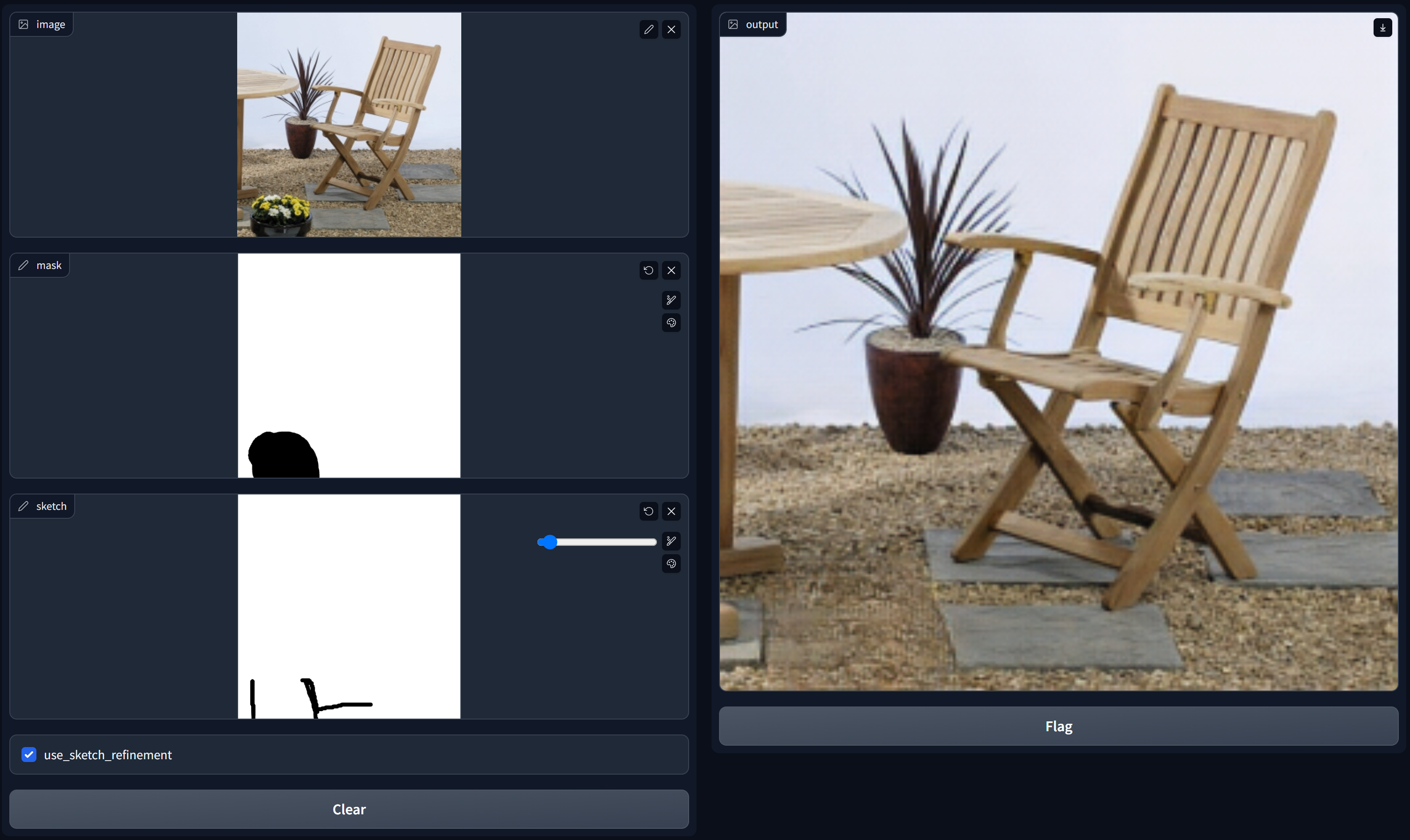} 
\caption{User interface of our designed interactive demo. Users are required to first upload an image for manipulation. Then, the sketching areas on the left side allow users to customize their desired mask and sketch interactively. Hereafter, users could turn on sketch refinement at any time by clicking the checkbox. Corresponding inpainted results would be simultaneously displayed on the right side. The \emph{Clear} button allows users to erase all their samples, and the \emph{Flag} button would save the uploaded image, customized mask and sketch, and corresponding result locally.}
\label{fig:demo}
\end{figure*}

\clearpage

\begin{table}[t]
    \centering
    \setlength{\tabcolsep}{0.8mm}
    \small
    \caption{Network architecture details of Registration Module (RM) in SRN. Note that there exists skip connection between the encoder and decoder of RM, where the first layer corresponds to the last layer, the second layer corresponds to the penultimate layer, and so on. GC: Gated convolution \cite{yu2018free}. IN: Instance Normalization \cite{DBLP:journals/corr/UlyanovVL16}.}
    \scalebox{0.9}{
    \begin{tabular}{cccccc}
     \toprule
     Convolution & Channels & Kernel Size & Stride & Activation & Normalization \\
     \midrule
     GC & 32 & 3 & 1 & LeakyReLU & IN \\
     GC & 64 & 3 & 2 & LeakyReLU & IN\\
     GC & 128 & 3 & 2 & LeakyReLU & IN\\
     GC & 192 & 3 & 2 & LeakyReLU & IN\\
     GC & 192 & 3 & 1 & LeakyReLU & IN\\
     GC & 192 & 3 & 1 & LeakyReLU & IN\\
     GC & 192 & 3 & 1 & LeakyReLU & IN\\
     \multicolumn{6}{c}{2x Nearest Upsampling} \\
     GC & 128 & 3 & 1 & LeakyReLU & IN\\
     \multicolumn{6}{c}{2x Nearest Upsampling} \\
     GC & 64 & 3 & 1 & LeakyReLU & IN\\
     \multicolumn{6}{c}{2x Nearest Upsampling} \\
     GC & 32 & 3 & 1 & LeakyReLU & IN\\
     GC & 1 & 3 & 1 & LeakyReLU & IN\\
     \bottomrule
    \end{tabular}
    }
    \label{network_rm}
\end{table}

\begin{table}[t!]
    \centering
    \setlength{\tabcolsep}{0.8mm}
    \small
    \caption{Network architecture details of Enhancement Module (EM) in SRN. VC: Vanilla Convolution. RCB: Residual Convolution Block, a simple convolution block with two stacks of vanilla convolution layers and residual connection. Note that there exists skip connection between the encoder and decoder of EM, where the first layer corresponds to the last layer, the second layer corresponds to the penultimate layer, and so on. ``-'' represents not using any normalization layer.}
    \scalebox{0.9}{
    \begin{tabular}{cccccc}
     \toprule
     Convolution & Channels & Kernel Size & Stride & Activation & Normalization \\
     \midrule
     \multicolumn{6}{c}{3 pixels Reflective Padding} \\
     VC & 32 & 7 & 1 & ReLU & - \\
     VC & 64 & 3 & 2 & ReLU & - \\
     VC & 128 & 3 & 2 & ReLU & - \\
     VC & 128 & 3 & 2 & ReLU & - \\
     VC & 128 & 3 & 1 & ReLU & - \\
     VC & 256 & 3 & 1 & ReLU & - \\
     RCB & 256 & 3 & 1 & ReLU & - \\
     RCB & 256 & 3 & 1 & ReLU & - \\
     RCB & 256 & 3 & 1 & ReLU & - \\
     RCB & 256 & 3 & 1 & ReLU & - \\
     VC & 128 & 3 & 1 & ReLU & - \\
     VC & 128 & 3 & 1 & ReLU & - \\
     \multicolumn{6}{c}{2x Nearest Upsampling} \\
     VC & 128 & 3 & 1 & ReLU & - \\
     \multicolumn{6}{c}{2x Nearest Upsampling} \\
     VC & 64 & 3 & 1 & ReLU & - \\
     \multicolumn{6}{c}{2x Nearest Upsampling} \\
     VC & 32 & 3 & 1 & ReLU & - \\
     VC & 1 & 3 & 1 & ReLU & - \\
     \bottomrule
    \end{tabular}
    }
    \label{network_em}
\end{table}

\begin{table}[t]
    \centering
    \setlength{\tabcolsep}{0.8mm}
    \small
    \caption{Network architecture details of Partial Sketch Encoder (PSE). GC: Gated convolution \cite{yu2018free}. BN: Batch Normalization \cite{ioffe2015batch}. SN: Spectral Normalization \cite{miyato2018spectral}. RCB: Residual Convolution Block, a simple convolution block with two stacks of vanilla convolution layers and residual connection. Note that ``-'' represents not using any activation layer.}
    \scalebox{0.9}{
    \begin{tabular}{cccccc}
     \toprule
     Convolution & Channels & Kernel Size & Stride & Activation & Normalization \\
     \midrule
     \multicolumn{6}{c}{3 pixels Reflective Padding} \\
     GC & 64 & 7 & 1 & ReLU & BN\\
     GC & 128 & 4 & 2 & - & BN\\
     GC & 256 & 4 & 2 & - & BN\\
     GC & 512 & 4 & 2 & - & BN\\
     RCB & 512 & 3 & 1 & ReLU & SN \& BN\\
     RCB & 512 & 3 & 1 & ReLU & SN \& BN\\
     RCB & 512 & 3 & 1 & ReLU & SN \& BN\\
     GC & 256 & 4 & 2 & - & BN\\
     GC & 128 & 4 & 2 & - & BN\\
     GC & 64 & 4 & 2 & - & BN\\
     \bottomrule
    \end{tabular}
    }
    \label{network_pse}
\end{table}

\begin{table}[t!]
    \centering
    \setlength{\tabcolsep}{0.8mm}
    \small
    \caption{Network architecture details of Texture Restoration Module (TRM). VC: Vanilla Convolution. TC: Transpose Convolution. BN: Batch Normalization \cite{ioffe2015batch}. SPADE: Spatially-Adaptive Normalization \cite{park2019semantic}. FFC: Fast Fourier Convolution \cite{suvorov2022resolution}. Note that ``-'' represents not using any activation layer or normalization layer.}
    \scalebox{0.9}{
    \begin{tabular}{cccccc}
     \toprule
     Convolution & Channels & Kernel Size & Stride & Activation & Normalization \\
     \midrule
     \multicolumn{6}{c}{3 pixels Reflective Padding} \\
     VC & 64 & 7 & 1 & ReLU & BN \\
     SFA & 128 & 3 & 2 & ReLU & SPADE \\
     SFA & 256 & 3 & 2 & ReLU & SPADE \\
     SFA & 512 & 3 & 2 & ReLU & SPADE \\
     FFC & 512 & 3 & 1 & ReLU & BN \\
     FFC & 512 & 3 & 1 & ReLU & BN \\
     FFC & 512 & 3 & 1 & ReLU & BN \\
     FFC & 512 & 3 & 1 & ReLU & BN \\
     FFC & 512 & 3 & 1 & ReLU & BN \\
     FFC & 512 & 3 & 1 & ReLU & BN \\
     FFC & 512 & 3 & 1 & ReLU & BN \\
     FFC & 512 & 3 & 1 & ReLU & BN \\
     FFC & 512 & 3 & 1 & ReLU & BN \\
     TC & 256 & 3 & 2 & - & BN\\
     TC & 128 & 3 & 2 & - & BN\\
     TC & 64 & 3 & 2 & - & BN\\
     \multicolumn{6}{c}{3 pixels Reflective Padding} \\
     VC & 3 & 7 & 1 & Tanh & - \\
     \bottomrule
    \end{tabular}
    }
    \label{network_trm}
\end{table}

\clearpage

\begin{figure*}[t!]
\centering
\includegraphics[width=1.0\linewidth]{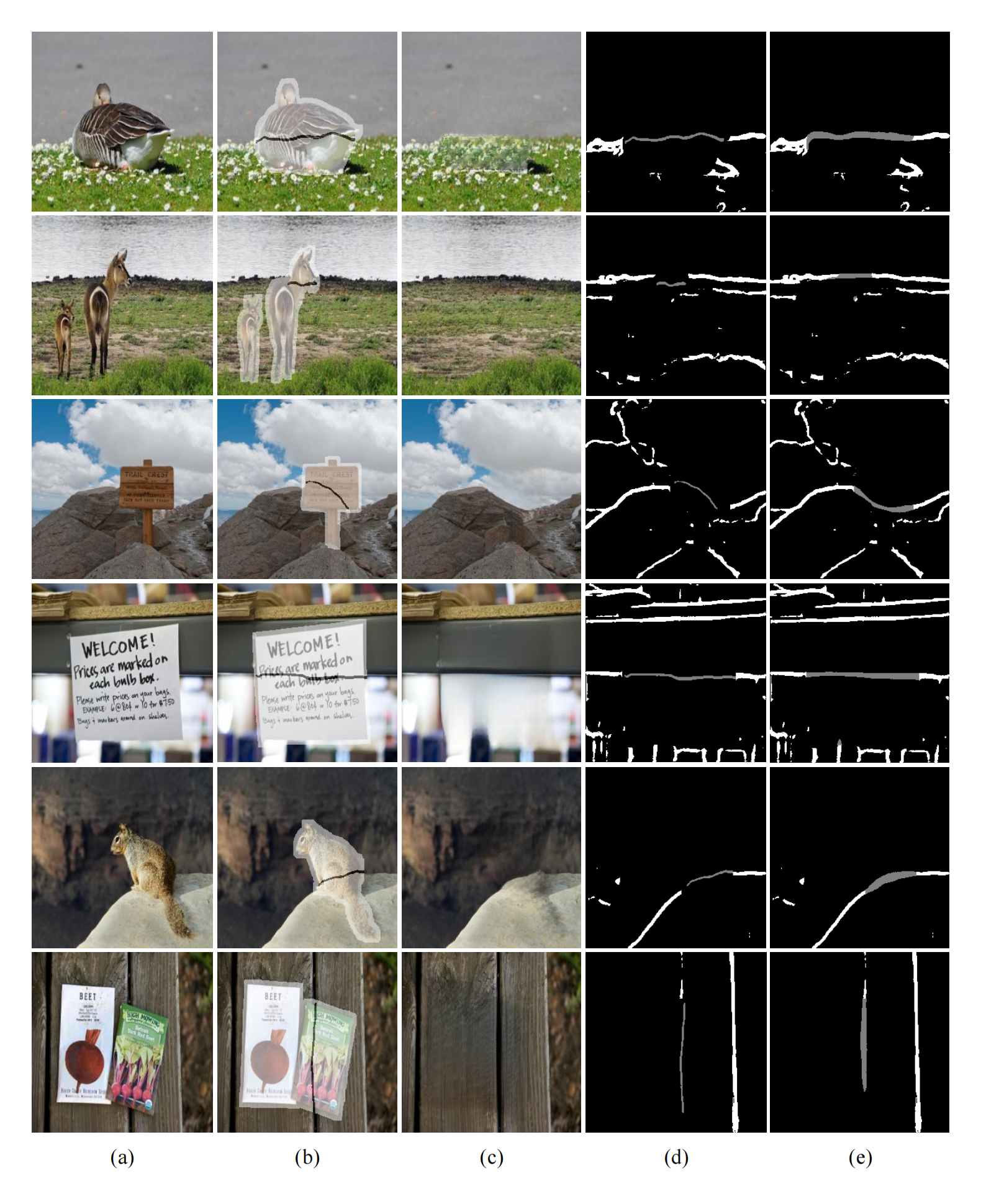} 
\caption{Visualization of refined sketches and corresponding inpainting results. (From left to right) (a) Image, (b) input, (c) result, (d) sketch, (e) refined sketch. Note that (b) represents the visualization of masked input with sketch. In (d) and (e), sketch strokes in \textbf{grey} represent strokes in the mask regions. \textbf{Zoom in for best view.}}
\label{fig2}
\end{figure*}

\begin{figure*}[t!]
\centering
\includegraphics[width=0.93\textwidth]{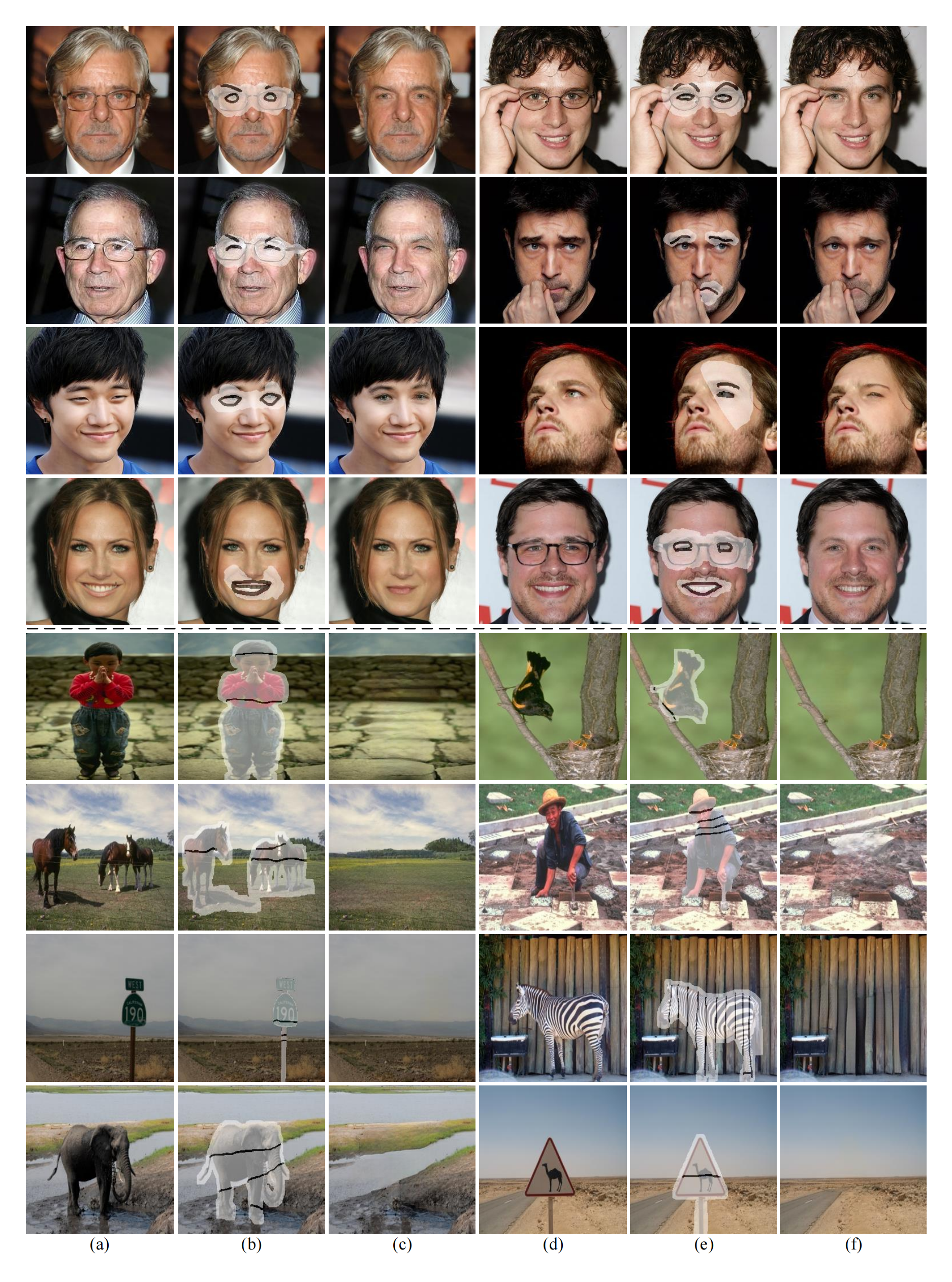} 
\caption{More qualitative results of face manipulation (top) and scene editing (bottom) upon our proposed test protocol. (From left to right) (a) Image, (b) input, (c) result, (d) image, (e) input, (f) result. Note that (b) and (e) represent visualizations of masked input with sketch. \textbf{Zoom in for best view.}}
\label{fig4}
\end{figure*}

\end{document}